\documentclass[journal]{IEEEtran}

\usepackage{times}
\usepackage[utf8]{inputenc}
\usepackage{epsfig}
\usepackage{graphicx}
\usepackage{xcolor,colortbl}
\usepackage[caption=false,font=footnotesize]{subfig}
\usepackage{amssymb}
\usepackage{amsmath}
\usepackage{pifont}
\usepackage{array}
\usepackage{bbold}
\usepackage{nicefrac}
\usepackage{tikz}
\usepackage{tikzscale}
\usepackage{pgfplots}
\pgfplotsset{compat=newest}
\usepackage[mode=buildnew]{standalone}
\usepackage{dsfont}
\usepackage{csquotes}
\usepackage{booktabs}
\usepackage{multirow}
\usepackage{siunitx}
\usepackage{diagbox}
\usepackage{bm}
\usepackage{enumitem}
\usepackage{algorithm} 
\usepackage{algpseudocode} 
\usepackage{hyperref}
\setitemize{leftmargin=*}

\newcommand{\stz}{\rule{0mm}{2.1ex}}

\newcommand{\btb}{     \begin{tabbing}             }
\newcommand{\bte}{     \end{tabbing}               }

\DeclareMathOperator*{\argmax}{argmax} 
\DeclareSymbolFont{AMSb}{U}{msb}{m}{n}
\DeclareSymbolFontAlphabet{\mathbb}{AMSb}

\newif\iftitsreview
\newif\ifshownumbers
\titsreviewtrue  % for the revision make all changes red
\titsreviewfalse % Uncomment this line for final version
\shownumberstrue % Show the numbers, to which the red comments belong to
\shownumbersfalse % Uncomment this line, to make numbers dissapear
\iftitsreview
\ifshownumbers
\newcommand{\revision}[2]{\textcolor{red}{#2 (#1)}}
\else
\newcommand{\revision}[2]{\textcolor{red}{#2}}
\fi
\else
\newcommand{\revision}[2]{#2}
\fi

\begin{document}

\title{Continual BatchNorm Adaptation (CBNA)\\ for Semantic Segmentation}

\author{Marvin Klingner \IEEEmembership{Student Member,~IEEE}, Mouadh Ayache, and Tim Fingscheidt, \IEEEmembership{Senior Member,~IEEE}
%\thanks{Manuscript received July 13, 2021\textcolor{red}{, revised February 17, 2022}}
\thanks{The authors are with the Institute of Communications Technology, Technische Universität Braunschweig, 38106 Braunschweig, Germany (e-mail: m.klingner@tu-bs.de; m.ayache@tu-bs.de; t.fingscheidt@tu-bs.de).}% <-this % stops a space
}

%\markboth{IEEE TRANSACTIONS ON INTELLIGENT TRANSPORTATION SYSTEMS}%
%{Klingner \MakeLowercase{\textit{et al.}}: Continual BatchNorm Adaptation}

\maketitle

\begin{abstract}
Environment perception in autonomous driving vehicles often heavily relies on deep neural networks (DNNs), which are subject to domain shifts, leading to a significantly decreased performance during DNN deployment. Usually, this problem is addressed by unsupervised domain adaptation (UDA) approaches trained \revision{4.1}{either simultaneously on source and target domain datasets or even source-free only on target data in an offline fashion. In this work, we further expand a source-free UDA approach to a continual and therefore online-capable UDA on a single-image basis for semantic segmentation. Accordingly, our method only requires the pre-trained model from the supplier (trained in the source domain) and the current (unlabeled target domain) camera image. Our method Continual BatchNorm Adaptation (CBNA) modifies} the source domain statistics in the batch normalization layers, using target domain images in an unsupervised fashion, which yields consistent performance improvements during inference. Thereby, in contrast to existing works, our approach can be applied to improve a DNN \revision{4.1}{continuously on a single-image basis} during deployment without access to source data, without algorithmic delay, and nearly without computational overhead. We show the consistent effectiveness of our method across a wide variety of source/target domain settings for semantic segmentation. Code is available at \href{https://github.com/ifnspaml/CBNA}{\url{https://github.com/ifnspaml/CBNA}}.
\end{abstract}


\IEEEpeerreviewmaketitle

\section{Introduction}
\label{sec:introduction}

\begin{figure}[t]
	\centering	
	\includestandalone[width=1.0\linewidth]{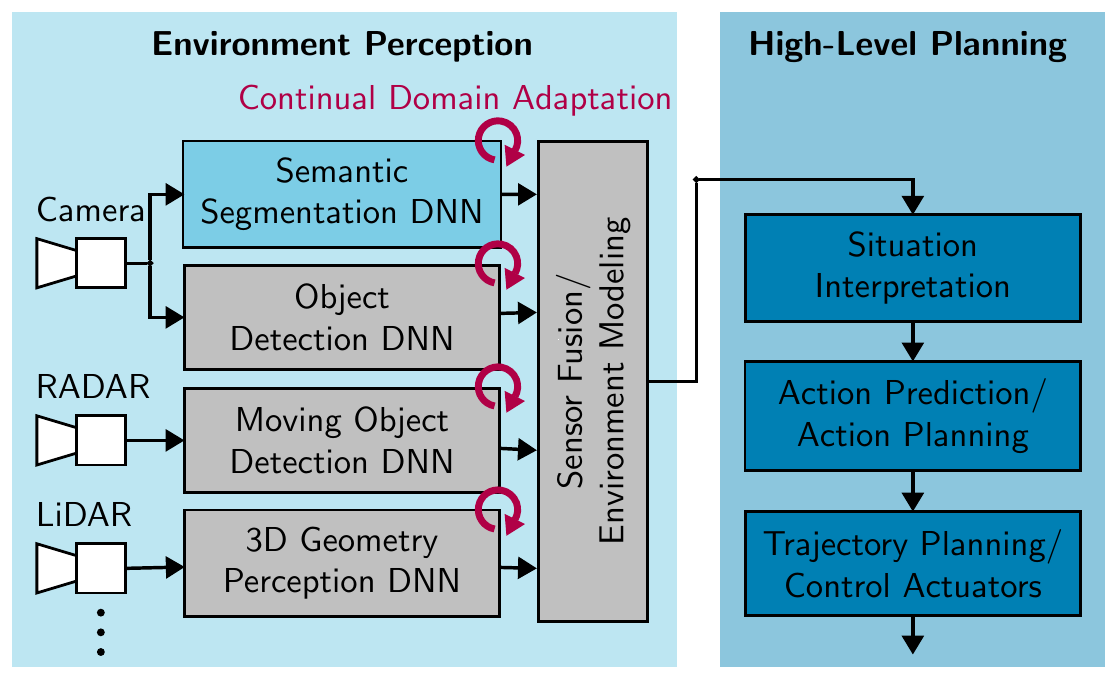}
	\caption{Overview about \textbf{continual unsupervised domain adaptation}, e.g., by CBNA, and its deployment in online environment perception.}
	\label{fig:top_level}
\end{figure}

The information processing concept of an autonomous driving vehicle as shown in Fig.~\ref{fig:top_level} relies heavily on deep neural networks (DNNs) \revision{1.3}{to extract information from sensor inputs such as camera images, RADAR measurements, or LiDAR point clouds. Exemplary tasks executed by such DNNs are semantic segmentation \cite{Long2015, Chen2018}, depth estimation \cite{Eigen2014, Zhou2017a}, instance segmentation \cite{He2016, Kirillov2019a}, or object detection \cite{Girshick2015, Chen2018b}, which are expected to provide high-quality outputs for a safe operation of the vehicle.} However, DNNs are usually trained on annotated datasets \cite{Geiger2013, Cordts2016}, only covering a small portion of real-life scenery. However, when DNNs are deployed in the car, the environment can change drastically due to, e.g., different image appearances from a new camera or day/night shifts, leading to a significantly decreased DNN performance \cite{Ganin2015, Tsai2018}. This problem \revision{4.2}{(known as domain shift \cite{Weiss2016})} needs to be addressed for a successful deployment of DNNs in highly automated vehicles.
\par
Focusing on the semantic segmentation task, two main concepts have been established to improve the performance in a real-world target domain that is unlabeled by nature. Firstly, in domain generalization (DG), the neural network is trained more robust on several different source domains to improve performance on unknown target domains \cite{Yue2019, Zhang2022}. Here, the target domain is assumed to be unavailable and accordingly one cannot make use of specific target domain images. Secondly, in unsupervised domain adaptation (UDA), the model is trained simultaneously on the labeled source data and unlabeled target data, assuming that data from both domains is available at the same time \cite{Tsai2018, Pan2020, Li2019b, Wang2020}. In practice, however, models are often trained on non-public datasets, which cannot be passed on due to data-privacy issues or for other practical reasons, meaning that neither source data nor representations thereof are available, instead only the trained model from the supplier is available for adaptation. In this case, DG as well as standard UDA techniques cannot be applied. Therefore, similar to \cite{Klingner2022}, we focus on UDA without source data, meaning that we adapt a given trained model using only unlabeled target domain data.
\par
\begin{figure}[t]
	\centering	
	\includestandalone[width=1.0\linewidth]{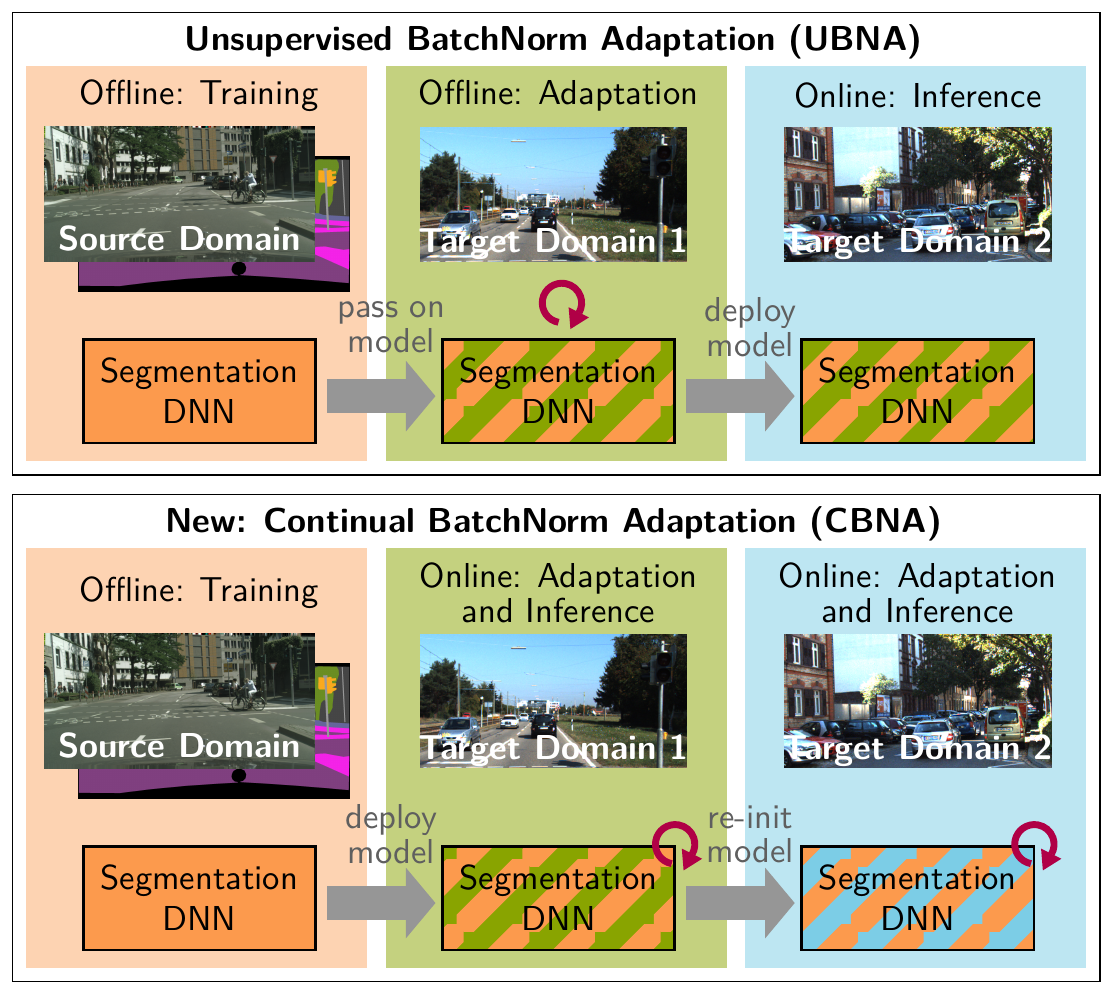}
	\caption{Overview on \textbf{how our novel CBNA approach differs from} other source-free UDA approaches, e.g., \textbf{UBNA} \cite{Klingner2022}. Re-initialization and adaptation with CBNA is performed for \textit{each} new image during inference.}
	\label{fig:concept_comparison}
\end{figure} 
In this work, we aim at a task which is even more challenging yet also more interesting for practical deployment: We focus on UDA \textit{without access to source data}, where the DNN is adapted \textit{during inference} for every single (target domain) image \textit{in a continual fashion}, see the bottom part of Fig.~\ref{fig:concept_comparison}. \revision{1.8}{Thereby, even if the domains switch rapidly in an image stream from a video (e.g., when driving into a tunnel), the network adapts to this on a single-image basis and can obtain optimal performance in each situation without any delay. Examples of such rapid domain changes could be a different camera illumination when driving into a tunnel or rapid weather/environmental changes, i.e., domain adaptation in adverse conditions \cite{Gong2021, Zhang2021}.} Previous standard UDA and DG approaches are of course inapplicable, as they require access to source data, which is usually unavailable in a highly automated vehicle due to storage limitations. Even our earlier Unsupervised BatchNorm Adaptation (UBNA) work \cite{Klingner2022} not relying on source data is not applicable to this task, as its adaptation takes place on a separate data subset from the target domain, see top part of Fig.~\ref{fig:concept_comparison}. If during deployment the domain changes again, or even permanently before inference can take place, the performance of the model decreases as the adaptation is not applied in a continual fashion. In practice, however, it would be desirable to adapt and infer the DNN on a single image basis at once to optimally match each new domain without algorithmic delay.
\par
To provide a solution for this defined task, we present our Continual BatchNorm Adaptation (CBNA) method (cf.~bottom part of Fig.~\ref{fig:concept_comparison}) as an extension of our previous work UBNA~\cite{Klingner2022}. Here, we \textit{mix the batch normalization (BN) statistics} (not the data!) of the source domain and a single target domain image in a continual fashion for each new image from the target domain. Thereby, during inference in a vehicle, we can adapt the deep neural network model instantaneously to each new image from the video stream of a camera, while previous approaches \cite{Klingner2022, Li2017b, Li2018d, Schneider2020, Zhang2022} adapt only once in an offline fashion to a single target domain using multiple uncorrelated images. Regarding computational complexity, CBNA introduces only little computational overhead on the forward pass during inference, while reference approaches reaching a similar performance \cite{Klingner2022, Schneider2020} would require a whole additional forward pass through the model. Note, that all aforementioned known approaches have been proposed for offline settings (cf. top part of Fig.~\ref{fig:concept_comparison}), making a direct comparison to our new CBNA proposal unfair due to the here envisaged more constrained continual UDA setting (cf. bottom part of Fig.~\ref{fig:concept_comparison}). Therefore, we report their performance in our framework under the same constraints as so-called \textit{reference methods}, as no baseline approaches for continual UDA of semantic segmentation exist so far.
\par
Our contributions with this work are as follows: Firstly, we present our online-capable CBNA method for continual UDA without source data of semantic segmentation models. Secondly, we show the successful applicability of CBNA on a single-image basis during inference, with only little computational overhead and no algorithmic delay being induced. Thirdly, we show the effectiveness of CBNA across a variety of source/target domain combinations, where we can even find hyperparameters which generalize across different target domains for a given segmentation model proving the practical applicability of CBNA. We will publish our code to facilitate further research on continual UDA without source data.
\par
This work is structured as follows. In Section~\ref{sec:related_work} we discuss related approaches. Afterwards, in Section~\ref{sec:bn_adaptation} we introduce our CBNA method as well as reference methods, followed by our experimental setup in Section~\ref{sec:experimental_setup}. We evaluate our method in Section~\ref{sec:experiments} and finally conclude this work in Section~\ref{sec:conclusion}.

\section{Related Work}
\label{sec:related_work}

We give an overview on related domain generalization (DG) and unsupervised domain adaptation (UDA) approaches. For UDA we particularly discuss approaches not relying on source data and approaches making use of normalization layers. 

\subsection{Domain Generalization (DG)}

The aim of DG methods \cite{Pan2018, Dou2019, Li2019c, Seo2020} is to improve DNN performance in an unknown target domain using data from (several different) source domains. For semantic segmentation, \revision{5.1}{several approaches have been proposed \cite{Pan2018, Yue2019, Choi2021}}, e.g., Yue~et~al.~\cite{Yue2019} mix the style of synthetic images with real images, using auxiliary source domain datasets, thereby learning more domain-invariant features. \revision{5.1}{While our CBNA method for continual source-free UDA is applied \textit{after} pre-training and using only the pre-trained model and \textit{target domain} data, DG is applied \textit{during} pre-training on \textit{source data} (usually with labels) and without target data. Thereby, if only a given trained model and unlabeled data from the target domain are available, DG methods cannot be applied, motivating the application of methods for continual UDA without source data.} Here, we additionally provide experimental results, where a DNN is first trained using DG methods and afterwards adapted using our CBNA algorithm for source-free continual UDA, showing that both methods for both tasks can be combined.

\subsection{Unsupervised Domain Adaptation (UDA)}

Standard UDA approaches assume that both labeled source data and unlabeled target data are available at the same time. Thereby, the domain transfer can be achieved in an offline fashion by using domain adaptation training techniques. These techniques can be roughly divided into three subcategories: Firstly,  \textit{domain-adversarial training} \cite{Bolte2019a, Chen2019c, Dong2020, Du2019, Hoffman2016, Huang2020, Tsai2018, Vu2019, Xu2019b}, can be applied, where domain-invariant features are learned by an additional discriminator (loss). Secondly, \textit{style transfer} \cite{Gong2019, Hoffman2018, Li2019b, Yang2020}, can be used to better match the appearance of source and target domain by image-to-image translation approaches. Thirdly, \textit{self-training} can be employed, where pseudo-labels are used as an additional supervision signal in the target domain \cite{Choi2019, Li2019b, Mei2020, Subhani2020, Zou2018}. As these approaches all require labeled source data to be available during the domain adaptation, they are not applicable when source data is not available, e.g., due to data privacy issues. If in this case an improvement in the target domain is still desired, UDA approaches not relying on source data have to be used instead.

\subsection{UDA Without Source Data}

Towards continual adaptation of semantic segmentation models, it is desirable to remove the need for source data during the adaptation, as this is usually a large dataset or a non-available dataset on the car manufacturer side, which cannot be stored on a deployed vehicle. The approaches of \cite{Kurmi2021, Wulfmeier2018} employ an auxiliary network which has been trained in the source domain together with the segmentation model. This network replays source domain knowledge to the network during adaptation. \revision{1.1}{Moreover, Stan~et~al.~\cite{Stan2020} and Termöhlen~et~al.~\cite{Termoehlen2021} learn a source domain distribution, which is aligned with the target domain distribution during adaptation.} These approaches do not make use of source data during the UDA. However, they still require an additional source domain representation for their approach (e.g., an additional network), which is usually also not available for a trained model.
\par
\revision{1.1}{Only few approaches exist for UDA of a \textit{given trained} model relying only on unlabeled target domain data. Some initial approaches relying on training with pseudo labels \cite{Li2020c, Yeh2021}, alignment methods for the latent space distribution \cite{Li2020, Liang2020, Yeh2021}, or class-conditional generative adversarial networks \cite{Li2020a} focus on simple tasks such as image classification or object detection. However, the aforementioned approaches do not address the semantic segmentation task, which we address in this work. For this task some very recent methods have been developed concurrently: Teja~et~al.~\cite{Teja2021} apply entropy minimization on the posterior and maximize the noise robustness of latent features. Kundu~et~al.~\cite{Kundu2021} use self-training on pseudo labels. Liu~et~al.~\cite{Liu2021} also make use of this technique and in addition apply data-free knowledge distillation. Our main distinguishing aspect from these works is the proposal of an efficient continual domain adaptation on a single-image basis, while to the best of our knowledge all other source-free methods for semantic segmentation rely on a time-consuming second training stage on many images in the target domain.}

\subsection{UDA via Normalization Layers}

The initial works of Li~et~al.~\cite{Li2017b, Li2018d} for image classification and Zhang~et~al.~\cite{Zhang2022} for semantic segmentation show that the re-estimation of batch normalization (BN) statistics in the target domain can be used for UDA without source data. The UBNA method from Klingner~et~al.~\cite{Klingner2022} has shown that mixing statistics from the source and target domain outperforms these initial works, which we build upon for our method design. These findings for domain adaptation are also supported by the work on adversarial robustness of Schneider~et~al.~\cite{Schneider2020}, where the beneficial effect of mixing statistics from perturbed and clean images is shown. However, the approaches mentioned before are only applicable to an offline UDA on a dataset, i.e., they still require statistics from multiple uncorrelated images in the target domain for a successful application. This is disadvantageous for continual UDA settings, where it would be desirable to continuously adapt on a single-image basis to avoid algorithmic delay during deployment in rapidly changing domains. In contrast to existing methods \cite{Klingner2022, Li2017b, Li2018d, Schneider2020, Zhang2022}, our CBNA method is applicable to these continual UDA settings, which we will show by our successful single-image adaptation results without the usage of additional uncorrelated images. Another novelty of CBNA is its integration into the single-image inference forward pass of an already trained model, which introduces nearly no computational overhead during inference.

\section{BatchNorm Adaptation Methods}
\label{sec:bn_adaptation}

In this section we first revisit the batch normalization (BN) layer and thereby introduce notations. Afterwards, we provide reference methods for continual UDA of BN parameters during inference, which we derive from their originally published offline versions. Finally, we introduce our novel CBNA method.

\subsection{Revisiting the Batch Normalization Layer, Notations}

As our adaptation method relies on the usage of batch normalization (BN) layers, we briefly revisit the BN operation for the scope of a fully convolutional DNN with two spatial dimensions following \cite{Ioffe2015}. Each BN layer then processes a batch of input feature maps $\bm{f}_\ell\in\mathbb{R}^{B\times H_\ell\times W_\ell\times C_\ell}$ with batch size $B$, height $H_\ell$, width $W_\ell$, and number of channels $C_\ell$ of the feature map in the BN layer with index $\ell$. Then the normalization is given by
\begin{equation}
	\hat{f}_{b, \ell, i, c} = \gamma_{\ell, c} \cdot \left(f_{b, \ell, i, c} - \mu_{\ell, c}\right)\cdot\left(\sigma_{\ell, c}^2 + \epsilon\right)^{-\frac{1}{2}} + \beta_{\ell, c}\;,
	\label{eq:batch_norm}
\end{equation}
where each feature $f_{b, \ell, i, c} \in \mathbb{R}$ is normalized over the batch and spatial dimensions with indices $b \in \mathcal{B} = \left\lbrace 1, ... , B \right\rbrace$ and $i \in \mathcal{I}_\ell = \left\lbrace 1, ... , H_\ell \cdot W_\ell \right\rbrace$, respectively, on a channel-wise basis (channel index $c \in \mathcal{C}_\ell = \left\lbrace 1, ... , C_\ell \right\rbrace$), yielding the normalized output $\hat{\bm{f}}_\ell$. In (\ref{eq:batch_norm}), $\bm{\mu}_\ell = (\mu_{\ell, c}) \in \mathbb{R}^{C_\ell}$ and $\bm{\sigma}_\ell = (\sigma_{\ell, c}) \in \mathbb{R}_{+}^{C_\ell}$ are the channel-wise computed mean and standard deviations in layer $\ell$, respectively, while $\bm{\gamma}_\ell = (\gamma_{\ell, c}) \in \mathbb{R}^{C_\ell}$ and $\bm{\beta}_\ell = (\beta_{\ell, c}) \in \mathbb{R}^{C_\ell}$ are learnable scaling and shifting parameters. The constant $\epsilon > 0$ is a small number avoiding divisions by zero.
\par
During learning step $k$ in training, the mean vector $\hat{\bm{\mu}}^{(k)}_\ell=(\hat{\mu}^{(k)}_{\ell, c})$ and standard deviation vector $\hat{\bm{\sigma}}^{(k)}_\ell=(\hat{\sigma}^{(k)}_{\ell, c})$ of the features $\bm{f}_{\ell}$ from the current batch $\mathcal{B}$ are calculated as 
\begin{align}
	\hat{\mu}^{(k)}_{\ell, c} &= \frac{1}{B H_\ell W_\ell}\sum_{b\in \mathcal{B}} \sum_{i\in \mathcal{I}_{\ell}} f_{b, \ell, i, c},
	\label{eq:mean}\\
	\left(\hat{\sigma}^{(k)}_{\ell, c}\right)^2 &= \frac{1}{B H_\ell W_\ell}\sum_{b\in \mathcal{B}} \sum_{i\in \mathcal{I}_{\ell}} \left( f_{b, \ell, i, c} - \hat{\mu}^{(k)}_{\ell, c}\right)^2.
	\label{eq:variance}
\end{align}
During training, these values are directly used for the forward pass computation in (\ref{eq:batch_norm}), i.e., $\bm{\mu}_{\ell} = \hat{\bm{\mu}}^{(k)}_{\ell}$ and $\bm{\sigma}_{\ell} = \hat{\bm{\sigma}}^{(k)}_{\ell}$. However, during inference, one does not desire a normalization over the batch dimension, as this would make the output of the DNN on one image dependent on the other images in the batch, inducing indeterministic performance. Therefore, as preparation for inference, the BN statistics of the entire training dataset is approximated by recursively tracking mean and variance from (\ref{eq:mean}) and (\ref{eq:variance}) as  
\begin{align}
  \check{\mu}_{\ell,c}^{(k)} &= \left( 1- \eta \right)\cdot \check{\mu}_{\ell,c}^{(k-1)} + \eta \cdot \hat{\mu}_{\ell,c}^{(k)}, \label{eq:running_mean}\\
  \left(\check{\sigma}_{\ell,c}^{(k)}\right)^2 &= \left( 1- \eta \right)\cdot \left(\check{\sigma}_{\ell,c}^{(k-1)}\right)^2 + \eta \cdot \left(\hat{\sigma}_{\ell,c}^{(k)}\right)^2, \label{eq:running_variance}
\end{align}
using a momentum parameter $\eta\in \left[0,1\right]$. The final values from (\ref{eq:running_mean}) and (\ref{eq:running_variance}) after $K$ learning steps are then stored and used later for inference, i.e., in (\ref{eq:batch_norm}) we employ $\bm{\mu}_{\ell} = \check{\bm{\mu}}^{(K)}_{\ell}$ and $\bm{\sigma}_{\ell} = \check{\bm{\sigma}}^{(K)}_{\ell}$. 

\subsection{Continuous Adaptation Reference Methods (\textbf{C-X})}
\label{sec:reference_methods}

To improve the semantic segmentation DNN's performance during inference, we aim at adapting to each single image $\bm{x}^{\mathcal{D}^{\mathrm{T}}}_t$ from the target domain $\mathcal{D}^{\mathrm{T}}$ from a video at time $t$, implementing a continual UDA. We assume that besides the input image $\bm{x}^{\mathcal{D}^{\mathrm{T}}}_t$ only the trained model parameters from the source domain are available for this purpose. For semantic segmentation, there are no baseline methods known for this task, however, we still want to allow a comparison to previous works and therefore we modify several approaches to fit into our defined task, which then serve as reference approaches \textbf{C-X} to our CBNA method. 
\par
The first such reference is a version of the AdaBN approach from Li~et~al.~\cite{Li2017b, Li2018d}, who replace the source domain's BN statistics $\check{\bm{\mu}}_{\ell}^{(K)}$, $\check{\bm{\sigma}}_{\ell}^{(K)}$ by the target domain's BN statistics during inference. Originally, Li~et~al. employ the statistics from all uncorrelated images of the test set in the computation. This, however, is not suitable for our single-image continual UDA task and would incur a large algorithmic delay of the method. Therefore, to fit into our task definition, we modify AdaBN ~\cite{Li2017b, Li2018d} as follows: In (\ref{eq:batch_norm}), the BN mean $\mu_{\ell, c}$ and variance $\sigma^2_{\ell, c}$ of each layer $\ell$ and channel $c$ are set individually for each target domain image $\bm{x}^{\mathcal{D}^{\mathrm{T}}}_t$ during inference, i.e., $\mu_{\ell, c}\equiv\mu_{t, \ell, c}$ and $\sigma^2_{\ell, c}\equiv\sigma^2_{t, \ell, c}$. They are calculated as
\begin{equation}
    \mu_{t, \ell, c} = \hat{\mu}_{t,\ell,c}\;\; \text{and}\;\; \sigma^2_{t,\ell, c} = \hat{\sigma}^2_{t,\ell, c},
    \label{eq:adabn}
\end{equation}
where $\hat{\mu}_{t,\ell,c}$ and $\hat{\sigma}^2_{t,\ell, c}$ are computed according to (\ref{eq:mean}) and (\ref{eq:variance}), respectively, using only a batch size of $B=1$, which is only the available single-image input $\bm{x}^{\mathcal{D}^{\mathrm{T}}}_t$. We dub this method \textbf{C-Li}, ``continuous Li'', noting that this procedure requires only a single forward pass during inference, as (\ref{eq:adabn}) can be computed during the inference forward pass.\footnote{This is essentially the same (efficient) computation which is also carried out during training of the BN layer with a batch size of $B=1$.} Interestingly, the approach from Zhang~et~al.~\cite{Zhang2022} reduces to the same formulation, if only a single target-domain image is used for adaptation during inference. We dub it \textbf{C-Zhang}.
\par
The second reference method is derived from the UBNA approach of Klingner~et~al.~\cite{Klingner2022} (\textbf{C-Klingner}), which adapts a model on a separate adaptation set by mixing the source domain BN statistics with the target domain BN statistics. While they do this using 50 adaptation steps, a separate adaptation to each single image with 50 additional forward passes may cause too much computational overhead for deployment of the method in a vehicle. However, it can be shown that in the limit of using the same single adaptation image in all 50 adaptation steps, UBNA can be reduced to a single \textit{additional} forward pass. On the first forward pass, the statistics $\hat{\bm{\mu}}_{t,\ell}$, $\hat{\bm{\sigma}}_{t,\ell}$ of the target domain image $\bm{x}^{\mathcal{D}^{\mathrm{T}}}_t$ are determined as in C-Li. Afterwards, before the second forward pass with the same image, the image-specific BN statistics $\bm{\mu}_{t, \ell}$, $\bm{\sigma}_{t, \ell}$ used during inference are updated element-wise as:
\begin{align}
  \mu_{t,\ell,c} &= \left( 1- \eta^{\mathcal{D}^{\mathrm{S}}} \right)\cdot \check{\mu}_{\ell,c}^{(K)} + \eta^{\mathcal{D}^{\mathrm{S}}} \cdot \hat{\mu}_{t,\ell,c}, \label{eq:ubna_mean}\\
  \sigma_{t,\ell,c}^2 &= \left( 1- \eta^{\mathcal{D}^{\mathrm{S}}} \right)\cdot \left(\check{\sigma}_{\ell,c}^{(K)}\right)^2 + \eta^{\mathcal{D}^{\mathrm{S}}} \cdot \hat{\sigma}_{t,\ell,c}^2, \label{eq:ubna_variance}
\end{align}
by additionally considering the source-domain statistics $\check{\bm{\mu}}_{\ell}^{(K)}$, $\check{\bm{\sigma}}_{\ell}^{(K)}$ (obtained from (\ref{eq:running_mean}) and (\ref{eq:running_variance}) after $K$ training steps), which were disregarded in C-Li. The mixing weight $\eta^{\mathcal{D}^{\mathrm{S}}}\in \left[0,1\right]$ is used to weigh the influence of the target domain statistics. Interestingly, the same formulation can be derived from the method of \cite{Schneider2020}, although they use a different hyperparameter formulation for $\eta^{\mathcal{D}^{\mathrm{S}}}$ and apply their method to improve adversarial robustness. While the mixing of source and target-domain statistics in C-Klingner \revision{4.3}{by (\ref{eq:ubna_mean}) and (\ref{eq:ubna_variance})} is shown to be beneficial for performance, it also induces a second forward pass, which is disadvantageous in terms of computational complexity.

\subsection{Novel Continuous BatchNorm Adaptation (CBNA)}

\begin{algorithm}[t]
	\caption{Model adaptation and inference with \textbf{CBNA}} 
	\begin{algorithmic}[1]
	    \State Load segmentation model trained on source data, \quad\quad\quad\phantom{}
		including the source domain's BN statistics $\check{\bm{\mu}}^{(K)}_{\ell}$, $\check{\bm{\sigma}}^{(K)}_{\ell}$ as trained in $K$ steps of (\ref{eq:running_mean}), (\ref{eq:running_variance})
		\State Take current image $\bm{x}^{\mathcal{D}^{\mathrm{T}}}_t$ from the target domain $\mathcal{D}^{\mathrm{T}}$
		\State \textbf{CBNA}: Initialize BN momentum $\eta^{\mathcal{D}^{\mathrm{S}}}$ for all BN layers $\ell$
		\State Pass image $\bm{x}^{\mathcal{D}^{\mathrm{T}}}_t$ through the model until the first BN layer
		\For {BN layer $\ell\in\left\lbrace 1,\ldots,L\right\rbrace$}
            \State \textbf{CBNA}: Calculate BN statistics according to (\ref{eq:cbna_feature_mean}), (\ref{eq:cbna_feature_variance})
			\State \textbf{CBNA}: Update BN statistics according to (\ref{eq:cbna_mean}), (\ref{eq:cbna_variance})
			\State Pass features through the BN layer according to (\ref{eq:batch_norm})
			\State Pass features further until the next BN layer $\ell + 1$
		\EndFor
		\State Pass features up to the end and generate the output $\bm{y}^{\mathcal{D}^{\mathrm{T}}}_t$
	\end{algorithmic} 
	\label{alg:cbna}
\end{algorithm}

\begin{figure}[t]
	\centering
	\includestandalone[width=1.0\linewidth]{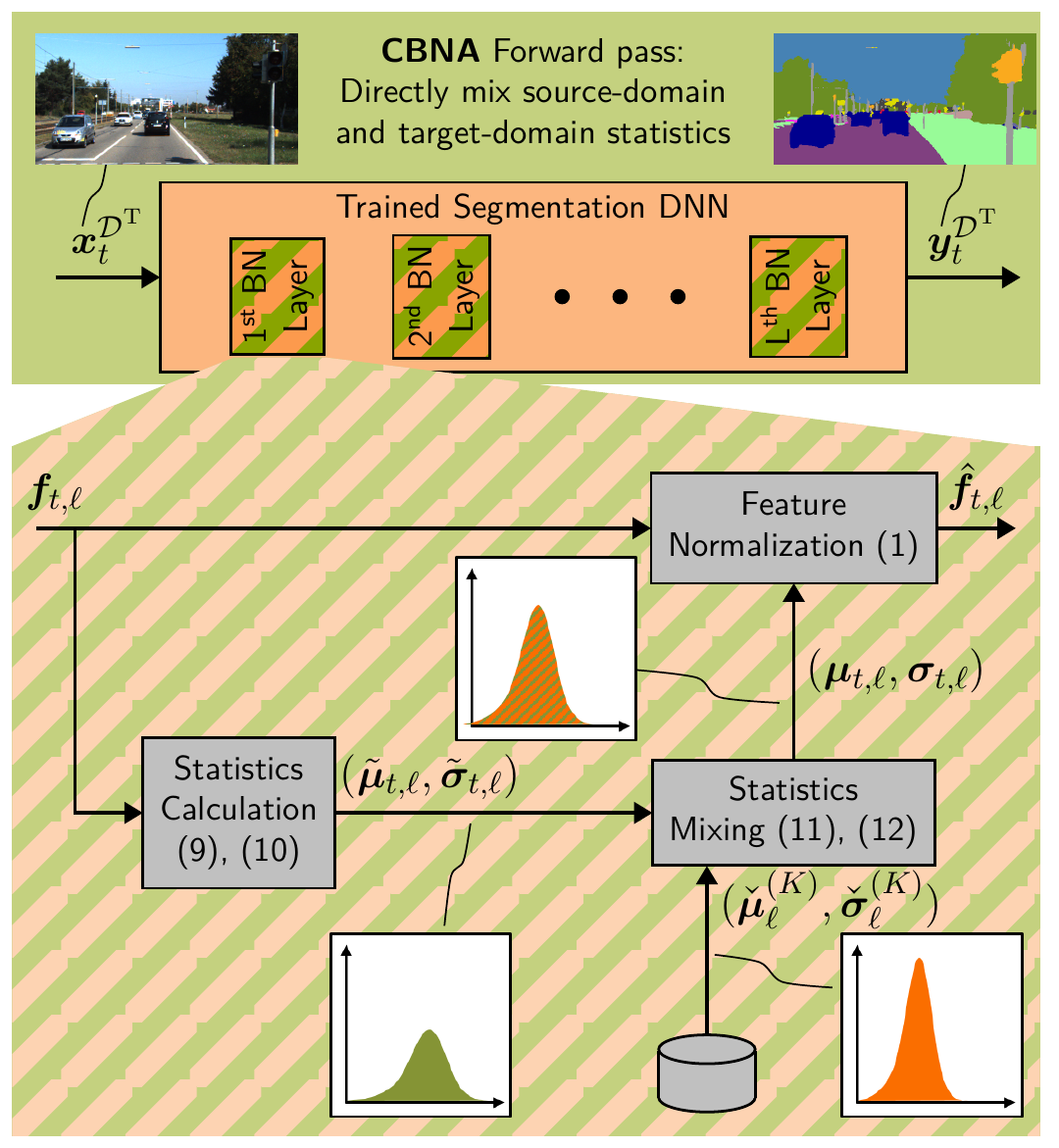}
	\caption{\revision{1.2}{Overview on how our \textbf{novel CBNA} approach mixes source and target domain BN statistics on a single forward pass. The color code shows, whether the network parts  are optimized using information from the target-domain image (green) or the source-domain data (orange).}}
	\label{fig:method_comparison}
\end{figure} 

\revision{1.2}{While both presented reference methods C-X come with the mentioned disadvantages, our \textbf{CBNA} method is able to mix BN statistics $\check{\bm{\mu}}^{(K)}_{\ell}$, $\check{\bm{\sigma}}^{(K)}_{\ell}$ from the source domain and the statistics of a single target-domain image $\bm{x}^{\mathcal{D}^{\mathrm{T}}}_t$ during a \textit{single} inference forward pass as shown in Fig.~\ref{fig:method_comparison}. Before the features $\bm{f}_{t,\ell}$ are normalized, their statistics are calculated and mixed with the stored source domain statistics. The mixed statistics are subsequently used to normalize the features. This method, also described by Algorithm \ref{alg:cbna}, is simply executed as one forward pass for each new target domain image $\bm{x}^{\mathcal{D}^{\mathrm{T}}}_t$.} \revision{3.1}{Note that in contrast to previous source-free UDA methods \cite{Teja2021, Kundu2021, Liu2021} our CBNA method is a \textit{continual} source-free UDA method (cf.~Fig.~\ref{fig:concept_comparison}).}
\par
\revision{5.3}{In contrast to C-Li, our CBNA method mixes source and target domain statistics, which is beneficial for performance and stability (cf.~Table~\ref{tab:baseline_comparison}). In contrast to C-Klingner, the mixing of source and target domain statistics is done in a single forward pass, which significantly reduces the additional computational complexity introduced through the continual adaptation (cf.~Table~\ref{tab:runtime_comparison}).}
\par
\revision{5.3}{The details of our proposed CBNA are as follows.} We initialize by imposing a weighting factor $\eta^{\mathcal{D}^{\mathrm{S}}}$ between source and target domain. This factor has to be chosen w.r.t.~the source domain model and should not differ for different target domains as the information about the target domain is only available during deployment and cannot always be known in advance. Notably, target domain information is, however, required for all previously proposed (offline) methods \cite{Li2017b, Li2018d, Zhang2022, Klingner2022}.
\par
During the \textit{single} inference forward pass, CBNA is applied, while the image $\bm{x}^{\mathcal{D}^{\mathrm{T}}}_t$ is processed by the segmentation DNN. When the feature processing in the DNN reaches BN layer $\ell$, we first compute the layer's image-specific BN statistics as 
\begin{align}
	\tilde{\mu}_{t,\ell, c} &= \frac{1}{H_\ell W_\ell} \sum_{i\in \mathcal{I}_{\ell}} f_{t,\ell, i, c}, \quad c\in\mathcal{C}_{\ell},
	\label{eq:cbna_feature_mean}\\
	\tilde{\sigma}_{t,\ell, c}^2 &= \frac{1}{H_\ell W_\ell} \sum_{i\in \mathcal{I}_{\ell}} \left( f_{t,\ell, i, c} - \tilde{\mu}_{t,\ell, c}\right)^2,\quad c\in\mathcal{C}_{\ell}, 
	\label{eq:cbna_feature_variance}
\end{align}
where the statistics are not only computed in dependency of the target domain image's statistics as in C-Li and C-Zhang and in the first forward pass of C-Klingner, but in dependency of the mixed statistics $\bm{\mu}_{t, \lambda}$ and $\bm{\sigma}_{t,\lambda}$ from all previous BN layers $\lambda \in\left\lbrace 1,2,...,\ell -1\right\rbrace$ (cf.~Fig~\ref{fig:method_comparison}), the $\ell$-th BN layer features depend upon\footnote{For the first layer ($\ell=1$), there is obviously no previous BN layer and therefore also no dependency of its statistics.}. Consequently, these image-specific statistics $\tilde{\bm{\mu}}_{t,\ell}$ and $\tilde{\bm{\sigma}}_{t,\ell}$ from (\ref{eq:cbna_feature_mean}) and (\ref{eq:cbna_feature_variance}), respectively, are \textit{applied immediately} to update the BN statistics used for normalization of the features $\bm{f}_{t,\ell}$ in BN layer $\ell$ as
\begin{align}
  \mu_{t,\ell,c} &= \left( 1- \eta^{\mathcal{D}^{\mathrm{S}}} \right)\cdot \check{\mu}_{\ell,c}^{(K)} + \eta^{\mathcal{D}^{\mathrm{S}}} \cdot \tilde{\mu}_{t,\ell, c},\quad c\in\mathcal{C}_{\ell}, \label{eq:cbna_mean}\\
  \sigma_{t,\ell,c}^2 &= \left( 1- \eta^{\mathcal{D}^{\mathrm{S}}} \right)\cdot \left(\check{\sigma}_{\ell,c}^{(K)}\right)^2 + \eta^{\mathcal{D}^{\mathrm{S}}} \cdot \tilde{\sigma}_{t,\ell, c} ^2,\quad c\in\mathcal{C}_{\ell}, \label{eq:cbna_variance}
\end{align}
Finally, the features are normalized according to (\ref{eq:batch_norm}) using the statistics from (\ref{eq:cbna_mean}) and (\ref{eq:cbna_variance}) and processed further until the next BN layer $\ell+1$. This procedure is repeated progressively through all BN layers of the model until the segmentation mask $\bm{y}^{\mathcal{D}^{\mathrm{T}}}_t$ has been generated (cf.~Algorithm~\ref{alg:cbna}). Note that the application of CBNA does only involve a single inference forward pass through the model with minimal computational overhead for computing (\ref{eq:cbna_feature_mean}) and (\ref{eq:cbna_feature_variance}) in each BN layer, and for updating the statistics in (\ref{eq:cbna_mean}) and (\ref{eq:cbna_variance}) in each BN layer, which presents a strong advantage over the C-Klingner reference method.
\par
\begin{figure}[t]
	\centering	
	\includestandalone[width=1.0\linewidth]{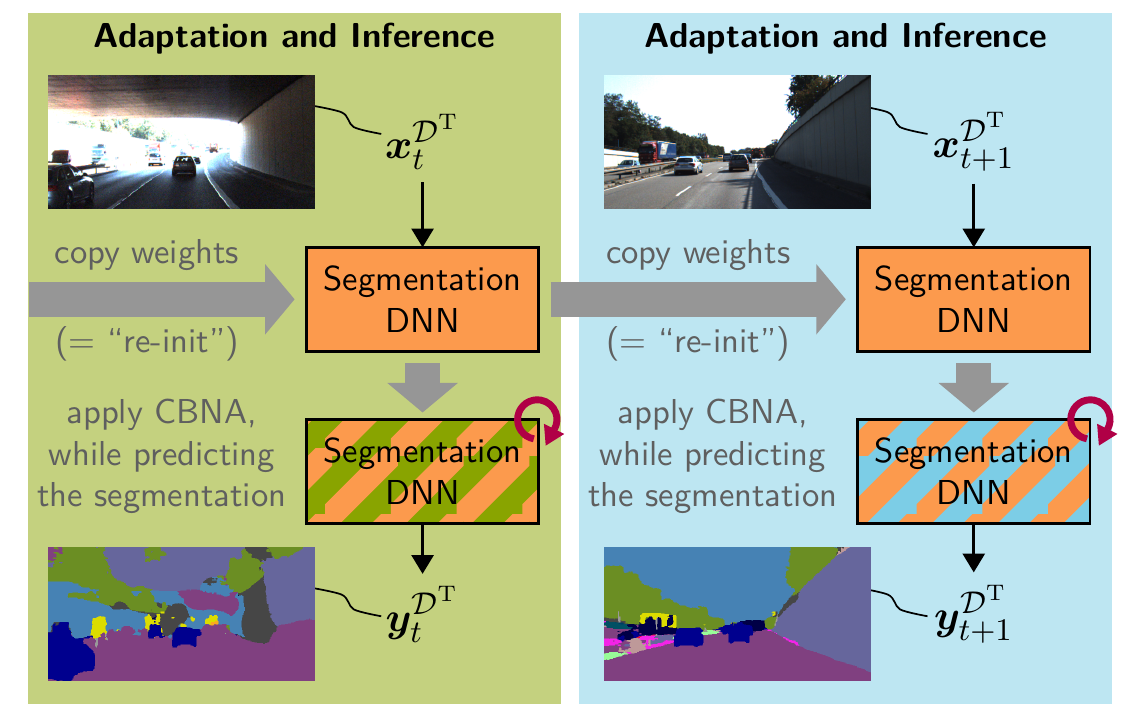}
	\caption{\textbf{Online adaptation and inference setup} of our \textbf{CBNA} method during deployment. Shown is a detail of Fig.~\ref{fig:concept_comparison} (bottom).}
	\label{fig:cbna_inference}
\end{figure} 
During deployment, our CBNA method can be used for DNN adaptation during the inference forward pass of each individual image $\bm{x}^{\mathcal{D}^{\mathrm{T}}}_t$ of a video as shown in Fig.~\ref{fig:cbna_inference}. Thereby, at time index $t$, the pre-trained model's BN statistics $\check{\bm{\mu}}^{(K)}_{\ell}$, $\check{\bm{\sigma}}^{(K)}_{\ell}$ are adapted to the current (target domain) image $\bm{x}^{\mathcal{D}^{\mathrm{T}}}_t$ by CBNA, as detailed in Algorithm~\ref{alg:cbna}. For the next image $\bm{x}^{\mathcal{D}^{\mathrm{T}}}_{t+1}$, again the pre-trained model from the source domain (with BN statistics $\check{\bm{\mu}}^{(K)}_{\ell}$, $\check{\bm{\sigma}}^{(K)}_{\ell}$) is used as re-initialization before the adaptation with CBNA during the inference forward pass. Thereby, CBNA can be applied in a continual fashion with very little computational overhead during deployment of a semantic segmentation DNN in a vehicle.

\section{Experimental and Evaluation Setup}
\label{sec:experimental_setup}

In this section, we first describe our used databases. Then, we explain our training procedures resulting in the given models for adaptation. Finally, we introduce evaluation metrics.

\subsection{Databases}

\begin{table}[t]
  \centering
  \setlength{\tabcolsep}{4pt}
  \caption{\textbf{Available databases} and their corresponding number of images used for training and for evaluation.}
  \begin{tabular}{l|c|rrr}
  %\hline
  \multirow{2}{*}{Dataset} & \multirow{2}{*}{Domain} & \multirow{2}{*}{pre-training} & adaptation\;\;\; & adaptation \\
   & & & \multicolumn{1}{c}{ \& validation} & \multicolumn{1}{c}{\& test}\\
  \hline
  \rule{0pt}{2.2ex}\!
  \!GTA-5 \cite{Richter2016} & $\mathcal{D}^{\mathrm{S}}$ & 24,966 & - & - \\
  SYNTHIA \cite{Ros2016} & $\mathcal{D}^{\mathrm{S}}$ & 9,400 & - & - \\
  Cityscapes \cite{Cordts2016} & $\mathcal{D}^{\mathrm{S}}$ & 2,975 & - & - \\
  KITTI \cite{Geiger2013, Menze2015} & $\mathcal{D}^{\mathrm{S}}$ & 200 & - & - \\
  \hline
  \rule{0pt}{2.2ex}\!
  \!Cityscapes \cite{Cordts2016} & $\mathcal{D}^{\mathrm{T}}$ & - & 500 & 2,975 \\
  KITTI \cite{Geiger2013, Menze2015} & $\mathcal{D}^{\mathrm{T}}$ & - & 200 & - \\
  BDD \cite{Yu2018b} & $\mathcal{D}^{\mathrm{T}}$ & - & 1,000 & - \\
  Mapillary \cite{Neuhold2017} & $\mathcal{D}^{\mathrm{T}}$ & - & 2,000 & - \\
  %\hline
  \end{tabular}
  \label{tab:dataset_overview}
\end{table}

We carry out experiments across a variety of datasets used for training the given models (top part of Tab.~\ref{tab:dataset_overview}) and for evaluation of the CBNA method (bottom part of Tab.~\ref{tab:dataset_overview}). Our main experiments use pre-trained models from the synthetic datasets GTA-5 ($\mathcal{D}^{\mathrm{S}}$) \cite{Richter2016} and SYNTHIA ($\mathcal{D}^{\mathrm{S}}$) \cite{Ros2016}, which are commonly used in other UDA works \cite{Tsai2018, Pan2020, Li2019b, Wang2020}. To show the applicability of CBNA to real-to-real adaptation settings we alternatively use the real dataset Cityscapes ($\mathcal{D}^{\mathrm{S}}$) \cite{Cordts2016} for training. The real dataset KITTI ($\mathcal{D}^{\mathrm{S}}$) \cite{Geiger2013} utilizing the 200 training images from the KITTI 2015 dataset \cite{Menze2015} is used as additional source-domain training material throughout the later described domain generalization (DG-Init) experiments.
\par
Although CBNA is meant to be applied to image sequences (i.e., a video) during deployment, there are no well-established video benchmarks for UDA of semantic segmentation. However, as CBNA and all C-X reference methods are applicable on a single-image basis, the evaluation can be carried out equivalently on single images of a validation/test set containing uncorrelated images. For our main experiments (based on GTA-5 and SYNTHIA training) we use the target domain Cityscapes ($\mathcal{D}^{\mathrm{T}}$) with 500 validation images to optimize our method's hyperparameters and 2,975 test images (official training images) to show their generalizability. Note that we use the official Cityscapes training images in our test set, as the official test set has no publicly available labels. Moreover, we use the target domains KITTI ($\mathcal{D}^{\mathrm{T}}$) \cite{Geiger2013, Menze2015}, BDD ($\mathcal{D}^{\mathrm{T}}$) \cite{Yu2018b}, and Mapillary ($\mathcal{D}^{\mathrm{T}}$) \cite{Neuhold2017} \revision{1.5}{(further details in Appendix \ref{sec:label_inconsistency})} during ablation experiments. Whenever the domains Cityscapes ($\mathcal{D}^{\mathrm{S}}$) or KITTI ($\mathcal{D}^{\mathrm{S}}$) have been used during training, we do not employ the respective target domains during evaluation.

\subsection{Training of the ``Given'' Source-Domain Models}

We use the same network architecture as in \cite{Klingner2022} relying on the widely used \texttt{VGG} \cite{Simonyan2015} and \texttt{ResNet} \cite{He2016} network architectures \revision{1.4}{(further details in Appendix \ref{sec:network_architecture})}. The input to the network is an RGB image $\bm{x}_t^{\mathcal{D}^{\mathrm{S}}}\in\mathbb{I}^{H\times W\times C}$ from the source domain $\mathcal{D}^{\mathrm{S}}$ with height $H$, width $W$, and number of channels $C=3$. The image is normalized to the range $\mathbb{I} = \left[0,1\right]$. The network predicts a posterior probability tensor $\bm{y}_t^{\mathcal{D}^{\mathrm{S}}} = (y^{\mathcal{D}^{\mathrm{S}}}_{t,i, s})\in\mathbb{I}^{H\times W\times |\mathcal{S}|}$, where $y^{\mathcal{D}^{\mathrm{S}}}_{t,i, s}$ is the probability that a pixel $\bm{x}^{\mathcal{D}^{\mathrm{S}}}_{t,i}\in\mathbb{I}^{C}$ with $i\in \mathcal{I} = \left\lbrace 1, ..., H\cdot W \right\rbrace$ belongs to class $s \in \mathcal{S} = \left\lbrace 1,...,|\mathcal{S}|\right\rbrace$. For simplicity, we henceforth set $\bm{x}_t^{\mathcal{D}^{\mathrm{S}}} = \bm{x}_t$ and $\bm{y}_t^{\mathcal{D}^{\mathrm{S}}} = \bm{y}_t$ in this section. During inference, the final class can be determined through $m_{t,i} = \argmax_{s \in \mathcal{S}} y_{t,i, s}$, yielding a pixel-wise semantic segmentation map $\bm{m}_t = (m_{t,i})\in \mathcal{S}^{H\times W}$. During training, the network is optimized using ground truth labels $\overline{\bm{m}}_t\in\mathcal{S}^{H\times W}$, which are one-hot encoded such that $\overline{m}_{t,i} = \argmax_{s \in \mathcal{S}} \overline{y}_{t,i, s}$, yielding a ground truth tensor $\overline{\bm{y}}_t = (\overline{y}_{t,i,s}) = \left\lbrace 0,1\right\rbrace^{H\times\ W\times |\mathcal{S}|}$. For optimization, we use the weighted cross-entropy loss 
\begin{equation}
J_t^{\mathrm{seg}} = -\frac{1}{H\cdot W}\sum_{i \in\mathcal{I}}\sum_{s \in\mathcal{S}} w_s \overline{y}_{t,i,s} \cdot \log\left(y_{t,i,s}\right), 
\label{eq:crossentropy_loss}
\end{equation}
where the class-wise weights $w_s$ are determined as in \cite{Paszke2016}.
\par

%\begin{figure}[t]
	%\centering	
	%\includestandalone[width=1.0\linewidth]{figs/architecture/architecture2}
	%\caption{\textbf{Training setup in the source domain using labeled data} resulting in the given pre-trained models. Continual UDA by CBNA or any other reference method C-X comes afterwards.}
	%\label{fig:architecture}
%\end{figure} 
During optimization \revision{5.3}{with (\ref{eq:crossentropy_loss}) as loss function}, we resize the images from GTA-5, SYNTHIA, Cityscapes, and KITTI to resolutions of $1024 \times 576$, $1024 \times 608$, $1024 \times 512$, and $1024 \times 320$, respectively. Subsequently, these resized images are randomly cropped to a resolution of $640 \times 192$. As data augmentations we use horizontal flipping, random brightness ($\pm 0.2$), contrast ($\pm 0.2$), saturation ($\pm 0.2$), and hue ($\pm 0.1$). We optimize our segmentation models for $20$ epochs ($10,000$ training steps approximately comprise one epoch), using the Adam optimizer~\cite{Kingma2015} and a batch size of $B=12$ if we only use a single dataset. As a simple DG method we use mixed batches from two datasets ($6$ images from each dataset), which we mark by (DG-Init) during evaluation. This may not be the latest state-of-the-art DG method, however, the scope of this work is not to optimize a DG method but merely to show that CBNA can be applied to given models that were trained by DG methods. The learning rate is initially set to $10^{-4}$ and reduced to $10^{-5}$ for the last $5$ epochs. 

\subsection{Evaluation Metrics}

We evaluate the semantic segmentation output by calculating the mean intersection-over-union (mIoU) \cite{Everingham2015}
\begin{equation}
	\mathrm{mIoU} = \frac{1}{|\mathcal{S}|}\sum_{s\in\mathcal{S}} \mathrm{IoU}_{s} = \frac{1}{|\mathcal{S}|}\sum_{s\in\mathcal{S}}\frac{\mathrm{TP}_{ s}}{\mathrm{TP}_{ s} + \mathrm{FP}_{s} + \mathrm{FN}_{s}}
	\label{eq:miou_offline}
\end{equation}
over all $|\mathcal{S}|=19$ classes as defined in \cite{Cordts2016}, except for models trained on SYNTHIA, where we follow common practices \cite{Lee2019d, Vu2019a} in evaluating over subsets of 13 and 16 classes. For each class $s$ the number of true positives ($\mathrm{TP}_{s}$), false negatives ($\mathrm{FN}_{s}$), and false positives ($\mathrm{FP}_{s}$), calculated between predictions $\bm{m}_t$ and ground truths $\overline{\bm{m}}_t$, are accumulated over all $T$ images of the validation/test set. 
% Accordingly, we can obtain a class-wise intersection-over-union performance $\mathrm{IoU}_{s}$, but not an image-wise performance. However, for online applications it would also be interesting to obtain the performance of a semantic segmentation DNN on a single-image basis. We therefore introduce an online mIoU metric 
% \begin{equation}
% 	\widetilde{\mathrm{mIoU}} = \frac{1}{T}\sum_{t\in\mathcal{T}}\mathrm{mIoU}_{t},
% 	\label{eq:miou_online}
% \end{equation}
% where $\mathcal{T} = \left\lbrace1, ..., T \right\rbrace$ is the set of image indices from the validation/test set. Here, the numbers of true positives ($\mathrm{TP}_{t, s}$), false negatives ($\mathrm{FN}_{t, s}$), and false positives ($\mathrm{FP}_{t, s}$) are computed image-wise to obtain an image-wise mean intersection-over-union $\mathrm{mIoU}_{t}$ according to (\ref{eq:miou_offline}). Afterwards, we average over all images of the test set according to (\ref{eq:miou_online}). 
For adaptation and evaluation \revision{5.3}{(with (\ref{eq:miou_offline}) being used as metric)} we resize the images from Cityscapes, KITTI, BDD, and Mapillary to resolutions of $1024 \times 512$, $1024 \times 320$, $1024 \times 576$, and $1024 \times 576$, respectively.

\section{Experimental Evaluation}
\label{sec:experiments}

To evaluate our method, we first give an ablation on how the hyperparameters of CBNA influence the method's performance. The final chosen model is then compared to several re-implemented reference methods (C-X) and to the only offline-capable UBNA from \cite{Klingner2022}. Finally, we compare our method on standard UDA benchmarks and give some qualitative results.

\subsection{CBNA Method Design and Ablation}
\label{sec:cbna_method_design}

\begin{figure}[t]
	\centering	
	\resizebox{\linewidth}{!}{\input{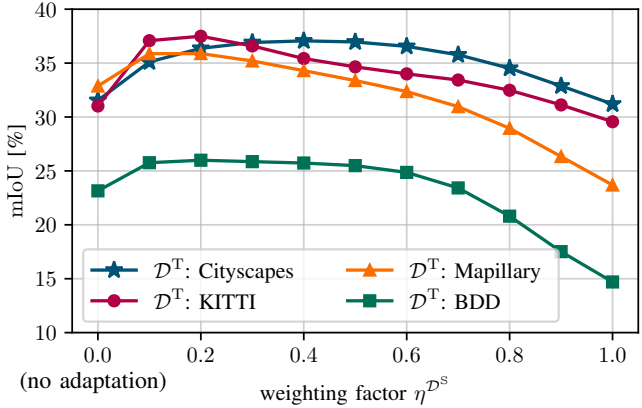}}
	\put(-240,12){\small (no adaptation)}
	\caption{\textbf{CBNA}: Influence of the weighting factor $\eta^{\mathcal{D}^{\mathrm{S}}}$ (\ref{eq:cbna_mean}), (\ref{eq:cbna_variance}) for the \texttt{VGG-16}-based model, when performing an adaptation from GTA-5 ($\mathcal{D}^{\mathrm{S}}$) to \textbf{several target domain validation sets} (cf.~Table~\ref{tab:dataset_overview}).}
	\label{fig:source_target_weighting}
\end{figure} 

When applying CBNA, first the question arises on how to weigh the influence of the source domain statistics and the statistics of the target domain image in (\ref{eq:cbna_mean}) and (\ref{eq:cbna_variance}), which is determined by the weighting factor $\eta^{\mathcal{D}^{\mathrm{S}}}$. The analysis in Fig.~\ref{fig:source_target_weighting} shows this influence for a \texttt{VGG-16}-based model being adapted from GTA-5 to several target domains, where $\eta^{\mathcal{D}^{\mathrm{S}}} = 0$ represents using only the source domain statistics (no adaptation), and $\eta^{\mathcal{D}^{\mathrm{S}}} = 1$ represents using only the target domain image's statistics (i.e., the C-Li method). We observe that the mIoU performance can be improved by approximately $3\%...5\%$ absolute (depending on the target domain) when mixing source and target domain statistics. However, if the influence from the target domain becomes too large, the performance decreases again. This is expected to some degree, as a high weight on the target domain image's BN statistics means that the network is strongly influenced by presumably rather unstable (i.e., highly time-variant) statistics of just a single image, compared to the statistics of many images from the source domain.
\par
For the considered source domain model in Fig.~\ref{fig:source_target_weighting}, different optimal weightings $\eta^{\mathcal{D}^{\mathrm{S}}}$ would be obtained for different target domains. However, in practice, we cannot assume prior knowledge about the target domain. Accordingly, each source domain model should only use a single weighting factor for all considered target domains and for any considered target image in general. Therefore, we choose the following strategy to obtain just a single weighting for all considered target domains: From the set $\mathcal{E}=\left\lbrace 0,0.1,0.2,0.3,0.4,0.5,0.6,0.7,0.8,0.9,1\right\rbrace$ we first take the best weighting for each target domain validation set. Then we average the target domain-specific weighting factors. Finally, we round to the next best weighting factor from $\mathcal{E}$. By this strategy we obtain a weighting factor of $\eta^{\mathcal{D}^{\mathrm{S}}}=0.2$ for models trained on GTA-5 and SYNTHIA, and $\eta^{\mathcal{D}^{\mathrm{S}}}=0.1$ for models trained on Cityscapes. In the following experiments we will use these weighting factors for all experiments. To ensure fairness, we optimize the reference method C-Klingner in the same fashion, while the reference methods C-Li and C-Zhang do not contain such hyperparameters and therefore do not need to be optimized.
%\begin{figure}[t]
%	\centering	
%	\resizebox{\linewidth}{!}{\input{figs/hyperparameters/different_windows.pgf}}
%	\caption{\textbf{CBNA}: Influence of considering additional preceding video frames ($\Delta N$ images in total) for the \texttt{VGG-16}-based model and an adaptation from \textbf{several source domains} to the Cityscapes ($\mathcal{D}^{\mathrm{T}}$) validation set.}
%	\label{fig:different_windows}
%\end{figure} 

\begin{figure}[t]
	\centering	
	\resizebox{\linewidth}{!}{\input{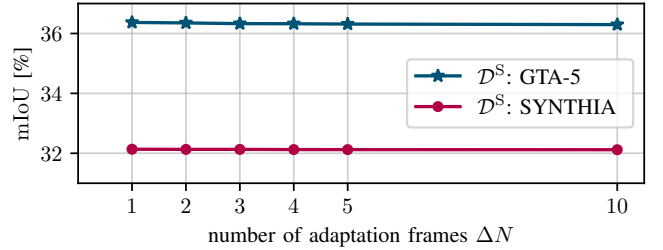}}
	\caption{\textbf{CBNA}: Influence of considering $\Delta N\! -\! 1$ preceding video frames for the \texttt{VGG-16}-based model, when performing an adaptation from \textbf{several source domains} to the Cityscapes ($\mathcal{D}^{\mathrm{T}}$) validation set.}
	\label{fig:different_windows}
\end{figure} 

\par
It is further of interest, whether the performance can be improved by considering the BN statistics from additional target domain images, which is investigated in Fig.\ \ref{fig:different_windows}. Here, for each sample, we additionally consider preceding image frames from its corresponding video (available in Cityscapes ($\mathcal{D}^{\mathrm{T}}$)). Interestingly, there is no gain in performance, indicating that the statistics of a single target domain image in combination with the source domain statistics is already sufficient for a stable adaptation. One could argue that this behavior could also be expected due to the high correlation between consecutive images. However, side experiments show that using random uncorrelated frames from the target domain yields essentially the same behavior as observed for highly correlated preceding video frames in Fig.~\ref{fig:different_windows}. Interestingly, this is not in contrast to \cite{Klingner2022}, where additional images improved the adaptation performance, as in \cite{Klingner2022} a single offline adaptation was used to adapt to the entire target domain (which can of course be better represented by statistics from several images), whereas CBNA adapts to each single image separately. Accordingly, we can draw the conclusion that for the scope of our method the adaptation can be done on a single image basis during inference, which is a huge advantage in terms of applicability and latency (reaction time to domain shift).

\subsection{Comparison to Reference Continual UDA Methods}

After having found suitable method hyperparameters, we facilitate a comparison to other possible approaches. As no continual UDA approaches for semantic segmentation exist so far, we reimplemented current related approaches and transferred them to our continual setting as described in Sec.~\ref{sec:reference_methods}. We compare the results to our CBNA method in Table~\ref{tab:baseline_comparison} for \texttt{VGG-16}-based and \texttt{ResNet-50}-based network architectures, with the adaptations from GTA-5 to Cityscapes and SYNTHIA to Cityscapes. The hyperparameters of CBNA and C-Klingner have been optimized for applicability to many target domains, as described in Section \ref{sec:cbna_method_design} on the validation sets of all target domains (cf.\ Table\ \ref{tab:dataset_overview}). We test their generalization to unseen data by utilizing the official Cityscapes training set as our test set. In Table~\ref{tab:baseline_comparison} we observe that just using the target domain statistics (C-Li and C-Zhang) does not improve the results significantly and even reduces the performance in some cases, e.g., for the adaptation of both architectures from SYNTHIA to Cityscapes. This is consistent with the observations from Fig.~\ref{fig:source_target_weighting}, where $\eta^{\mathcal{D}^{\mathrm{S}}} = 1$ decreased the performance in all cases. 
\begin{table}[t]
  \centering
  \setlength{\tabcolsep}{4pt}
  \caption{\textbf{Performance comparison of CBNA to re-simulated methods} modified to become continual single-image source-free UDA reference methods. Results are reported on \textbf{Cityscapes} ($\mathcal{D}^{\mathrm{T}}$).}
  \begin{tabular}{c|l|c|cc|cc}
  & & \\[-2.3ex]
  & Method & \rotatebox{90}{\shortstack{forward\\ passes}} & \multicolumn{2}{c|}{\shortstack{$\mathcal{D}^{\mathrm{S}}$:\,SYNTHIA\\ mIoU (\%)\\ (16 classes)}} &  \multicolumn{2}{c}{\shortstack{$\mathcal{D}^{\mathrm{S}}$:\,GTA-5\\ mIoU (\%)\\ (19 classes)}} \\
  \hline
   & & & validation & test & validation & test\\
  \hline
  \multirow{5}{*}{\rotatebox{90}{\texttt{ResNet-50}}} & No adaptation\stz & 1 & 29.2 & 30.0 & 33.6 &  35.0 \\
  \cline{2-7}
   & \textbf{C-Li}~($\sim$\cite{Li2018d})\stz & 1 & 28.8 & 28.9 &  34.3 & 35.6 \\ %AdaBN
   & \textbf{C-Zhang}~($\sim$\cite{Zhang2022}) & 1 & 28.8 & 28.9 & 34.3 & 35.6 \\ %TN
   & \textbf{C-Klingner}~($\sim$\cite{Klingner2022}) & 2 & \textbf{32.7} & \textbf{33.4} & \textbf{37.3} & \textbf{38.8} \\ %UBNA
  \cline{2-7}
   & \cellcolor[gray]{.90} & \cellcolor[gray]{.90} & \cellcolor[gray]{.90} & \cellcolor[gray]{.90} & \cellcolor[gray]{.90}  & \cellcolor[gray]{.90} \\[-2.3ex]
   & \cellcolor[gray]{.90} \!\!\textbf{CBNA} & \cellcolor[gray]{.90} \!\!1 & \cellcolor[gray]{.90} \!\!\underline{32.5} & \cellcolor[gray]{.90} \!\!\underline{33.2}  & \cellcolor[gray]{.90} \!\! \underline{36.7}  & \underline{38.3}     \cellcolor[gray]{.90} \\
  \hline
   \multirow{5}{*}{\rotatebox{90}{\texttt{VGG-16}}} & No adaptation\stz & 1 & 30.0  & 30.5 & 31.5 & 33.6 \\
  \cline{2-7}
   & \textbf{C-Li}~($\sim$\cite{Li2018d})\stz & 1 & 28.8  & 28.9 & 31.2 & 33.3 \\ %AdaBN
   & \textbf{C-Zhang}~($\sim$\cite{Zhang2022}) & 1 & 28.8 &  28.9 & 31.2 & 33.3 \\ %TN
   & \textbf{C-Klingner}~($\sim$\cite{Klingner2022}) & 2 & \textbf{33.4} & \textbf{33.7} & \textbf{36.7} & \textbf{39.0} \\ %UBNA
  \cline{2-7}
   & \cellcolor[gray]{.90}
   & \cellcolor[gray]{.90}  & \cellcolor[gray]{.90}  & \cellcolor[gray]{.90} & \cellcolor[gray]{.90}  & \cellcolor[gray]{.90} \\[-2.3ex]
   & \cellcolor[gray]{.90} \!\!\textbf{CBNA} & \cellcolor[gray]{.90} \!\!1 & \cellcolor[gray]{.90} \!\!\underline{32.1} & \cellcolor[gray]{.90} \!\!\underline{32.4} & \cellcolor[gray]{.90} \!\!\underline{36.4} & \cellcolor[gray]{.90} \!\!\underline{38.9} \\
  \end{tabular}
  \label{tab:baseline_comparison}
\end{table}

\par
The reference method C-Klingner, mixing source and target domain statistics, improves significantly over the ``no adaptation'' baseline, however, it involves the execution of a second forward pass, adding a lot of computations during inference (cf.\ Table\ \ref{tab:runtime_comparison}). A detailed analysis is given in the Appendix. In total, our CBNA method always performs second-ranked, close after C-Klingner, requiring only a single forward pass with almost no computational overhead. On the used \texttt{Tesla P100} graphics card, the \texttt{VGG-16}-based architecture with CBNA can be executed at \SI{20}{fps} (same as the ``no adaptation'' baseline), while C-Klingner reaches only \SI{10}{fps}. Compared to the source domain model, our CBNA method yields $3.2\%$ and $3.3\%$ absolute mIoU improvement for the adaptations from SYNTHIA to Cityscapes and GTA-5 to Cityscapes (test set), respectively, when applied to the \texttt{ResNet-50}-based model. Consistent improvements are also achieved for \texttt{VGG-16}-based models in these adaptation settings.

\subsection{Comparison to Offline Methods}

To better understand the advantages that our method offers, we also compare to the offline-capable UBNA method from \cite{Klingner2022} in Tables \ref{tab:cross_dataset_comparison_syn} and \ref{tab:cross_dataset_comparison_real}. Notably, UBNA is applied on 50 random images of the target domain, meaning that in order to apply this method in a vehicle, the time of domain switch needs to be known (otherwise complexity would be dramatically too high). In contrast, CBNA is applied on a single-image basis during inference, thus an adaptation to the current domain is applied in a continuous fashion. We therefore take a model from GTA-5 and adapt it to 4 different target domains with UBNA (cf.\ the right-hand side of Table~\ref{tab:cross_dataset_comparison_syn}). It is observable that UBNA improves the behavior on the domain it adapts to, e.g., from $33.6\%$ to $37.5\%$ for the \texttt{ResNet-50} model on Cityscapes. However, on other domains the performance often decreases, e.g., the \texttt{ResNet-50} model exhibits decreased performance on KITTI, BDD, and Mapillary when being adapted to Cityscapes. Similar behavior can also be observed for the other UBNA adaptations, when using a model pre-trained on GTA-5. In the same source domain condition (GTA-5), CBNA improves the performance for both \texttt{ResNet-50} and \texttt{VGG-16} in \textit{all} target domains. 
\begin{table}[t]
  \centering
  \setlength{\tabcolsep}{2pt}
  \caption{\textbf{Additional complexity} in 10$^9$ FLOPs/image for online adaptation in inference; image resolution of $512\times 1024$.}
  \begin{tabular}{l|c|ccc|c}
  Network & No adaptation & \textbf{C-Li} & \textbf{C-Zhang} & \textbf{C-Klingner} & \cellcolor[gray]{.90}\!\! \textbf{CBNA} \\
  \hline
  \texttt{ResNet-50} & 0 & $0.30$ & $0.30$ & $43$ & \cellcolor[gray]{.90}\!\! $0.30$ \\
  \texttt{VGG-16} & 0 & $0.43$ & $0.43$ & $161$ & \cellcolor[gray]{.90}\!\! $0.43$ \\
  \end{tabular}
  \label{tab:runtime_comparison}
\end{table}

\begin{table*}[t]
  \centering
  \setlength{\tabcolsep}{1.5pt}
  \caption{\textbf{Comparison to offline methods (UBNA \cite{Klingner2022})} across \textbf{various synthetic source domains and real target domain datasets} showing the strong online capability of our CBNA algorithm. mIoU values in \%; best results written in \textbf{boldface}.}
  \begin{tabular}{c|l|cccc|cccc}
  %\hline
  & & & \\[-2.3ex]
   & \multirow{2}{*}{Method} & \multicolumn{4}{c|}{\shortstack{$\mathcal{D}^{\mathrm{S}}$:\,SYNTHIA; mIoU (16 classes)}} &  \multicolumn{4}{c}{\shortstack{$\mathcal{D}^{\mathrm{S}}$:\,GTA-5; mIoU (19 classes)}} \\
  \cline{3-10}
  & & $\mathcal{D}^{\mathrm{T}}$:\,Cityscapes\stz & $\mathcal{D}^{\mathrm{T}}$:\,KITTI\stz & $\mathcal{D}^{\mathrm{T}}$:\,BDD\stz & $\mathcal{D}^{\mathrm{T}}$:\,Mapillary\stz & $\mathcal{D}^{\mathrm{T}}$:\,Cityscapes\stz & $\mathcal{D}^{\mathrm{T}}$:\,KITTI\stz & $\mathcal{D}^{\mathrm{T}}$:\,BDD\stz & $\mathcal{D}^{\mathrm{T}}$:\,Mapillary\stz \\
  \hline
  \multirow{6}{*}{\rotatebox{90}{\texttt{ResNet-50}}} & No adaptation\stz & 29.1 & \textbf{31.7} & 19.9 & \textbf{28.4} & 33.6 & \underline{36.9} & 30.0 & 34.4 \\
  \cline{2-10}
  & UBNA \cite{Klingner2022} (adapted to Cityscapes)\stz & \textbf{33.8} & 31.2 & 20.6 & 27.9 & \textbf{37.5} & 33.3 & 29.4 & 32.5 \\
  & UBNA \cite{Klingner2022} (adapted to KITTI) & 30.0 & 30.9 & 19.1 & 26.5 & 32.0 & 36.3 & 29.2 & 33.2 \\ 
  & UBNA \cite{Klingner2022} (adapted to BDD) &  31.1 & 30.9 & \underline{22.4} & 27.1 & \underline{37.2} & 32.2 & \textbf{32.8} & 36.3 \\ 
  & UBNA \cite{Klingner2022} (adapted to Mapillary) & 31.6 & \underline{31.6} & 20.3 & 26.8 & 36.6 & 33.3 & \underline{31.7} & \textbf{39.0}\\
  \cline{2-10}
  & \cellcolor[gray]{.90} & \cellcolor[gray]{.90} & \cellcolor[gray]{.90}& \cellcolor[gray]{.90} & \cellcolor[gray]{.90} & \cellcolor[gray]{.90} & \cellcolor[gray]{.90} & \cellcolor[gray]{.90} & \cellcolor[gray]{.90} \\[-2.3ex]
  & \cellcolor[gray]{.90} \!\!\textbf{CBNA} & \underline{32.5} \cellcolor[gray]{.90} & 31.2 \cellcolor[gray]{.90} & \textbf{22.6} \cellcolor[gray]{.90} & \underline{28.3} \cellcolor[gray]{.90} & 36.7 \cellcolor[gray]{.90} & \textbf{38.8} \cellcolor[gray]{.90} & 31.2 \cellcolor[gray]{.90} & \underline{37.8} \cellcolor[gray]{.90}\\
  %& \cellcolor[gray]{.90} \!\!\textbf{CBNA} (0.2) & 32.5 \cellcolor[gray]{.90} & 31.2 \cellcolor[gray]{.90} & 22.6 \cellcolor[gray]{.90} & 28.3 \cellcolor[gray]{.90} & 36.7 \cellcolor[gray]{.90} & 38.8 \cellcolor[gray]{.90} & 31.2 \cellcolor[gray]{.90} & 37.8 \cellcolor[gray]{.90}\\
  %& \cellcolor[gray]{.90} \!\!\textbf{CBNA} (0.3) & 32.1 \cellcolor[gray]{.90} & 29.9 \cellcolor[gray]{.90} & 22.2 \cellcolor[gray]{.90} & 26.2 \cellcolor[gray]{.90} & 36.6 \cellcolor[gray]{.90} & 36.3 \cellcolor[gray]{.90} & 30.7 \cellcolor[gray]{.90} & 36.4 \cellcolor[gray]{.90}\\
  %& \cellcolor[gray]{.90} \!\!\textbf{CBNA} (0.4) & 31.5 \cellcolor[gray]{.90} & 28.7 \cellcolor[gray]{.90} & 21.4 \cellcolor[gray]{.90} & 24.3 \cellcolor[gray]{.90} & 36.3 \cellcolor[gray]{.90} & 34.1 \cellcolor[gray]{.90} & 30.1 \cellcolor[gray]{.90} & 34.9 \cellcolor[gray]{.90}\\
 % & \cellcolor[gray]{.90} \!\!\textbf{CBNA}$^{\myplus}$ & \cellcolor[gray]{.90} & \cellcolor[gray]{.90} & \cellcolor[gray]{.90} & \cellcolor[gray]{.90} & \cellcolor[gray]{.90} & \cellcolor[gray]{.90} & \cellcolor[gray]{.90} & \cellcolor[gray]{.90} \\
  \hline
  \multirow{6}{*}{\rotatebox{90}{\texttt{VGG-16}}} & No adaptation\stz & 30.0 & 27.5 & 19.1 & \textbf{27.6} & 31.5 & 31.0 & 23.1 & 32.9 \\
  \cline{2-10}
  & UBNA \cite{Klingner2022} (adapted to Cityscapes)\stz & \textbf{34.4} & \textbf{29.5} & 17.4 & 26.5 & \underline{36.1} & 28.5 & 21.2 & 28.0\\
  & UBNA \cite{Klingner2022} (adapted to KITTI) & 30.3 & 28.9 & 15.8 & 25.2 & 31.5 & \underline{32.5} & 21.7 & 27.9 \\ 
  & UBNA \cite{Klingner2022} (adapted to BDD) &  \underline{32.3} & 27.7 & \textbf{19.7} & 26.1 & 33.6 & 26.3 & 25.0 & 30.8\\ 
  & UBNA \cite{Klingner2022} (adapted to Mapillary) & 31.1 & 28.5 & 17.7 & 26.2 & 33.9 & 29.9 & \underline{25.5} & \underline{34.8} \\
  \cline{2-10}
  & \cellcolor[gray]{.90} & \cellcolor[gray]{.90} & \cellcolor[gray]{.90} & \cellcolor[gray]{.90} & \cellcolor[gray]{.90} & \cellcolor[gray]{.90} & \cellcolor[gray]{.90} & \cellcolor[gray]{.90} & \cellcolor[gray]{.90} \\[-2.3ex]
  & \cellcolor[gray]{.90} \!\!\textbf{CBNA} & \cellcolor[gray]{.90} 32.1 &  \cellcolor[gray]{.90} \underline{29.4} & \cellcolor[gray]{.90} \underline{19.4} & \cellcolor[gray]{.90} \underline{27.0} & \cellcolor[gray]{.90} \textbf{36.4} &  \cellcolor[gray]{.90}  \textbf{37.5} & \cellcolor[gray]{.90} \textbf{26.0} & \cellcolor[gray]{.90}  \textbf{35.9} \\
  %& \cellcolor[gray]{.90} \!\!\textbf{CBNA (0.2)} & \cellcolor[gray]{.90} 32.1 &  \cellcolor[gray]{.90} 29.4 & \cellcolor[gray]{.90} 19.4 & \cellcolor[gray]{.90} 27.0 & \cellcolor[gray]{.90} 36.4 &  \cellcolor[gray]{.90}  37.5 & \cellcolor[gray]{.90} 26.0 & \cellcolor[gray]{.90}  35.9 \\
  %& \cellcolor[gray]{.90} \!\!\textbf{CBNA (0.3)} & \cellcolor[gray]{.90} 32.4 & \cellcolor[gray]{.90} 28.7 & \cellcolor[gray]{.90}  18.9 & \cellcolor[gray]{.90} 25.7 & \cellcolor[gray]{.90} 36.9 & \cellcolor[gray]{.90} 36.6 & \cellcolor[gray]{.90} 25.9 & \cellcolor[gray]{.90} 35.2 \\
  %& \cellcolor[gray]{.90} \!\!\textbf{CBNA (0.4)} & \cellcolor[gray]{.90} 32.3 &  \cellcolor[gray]{.90}  28.0 & \cellcolor[gray]{.90} 18.4 & \cellcolor[gray]{.90} 24.5 & \cellcolor[gray]{.90} 37.1 & \cellcolor[gray]{.90} 35.4 & \cellcolor[gray]{.90} 25.7 & \cellcolor[gray]{.90} 34.3\\
  %& \cellcolor[gray]{.90} \!\!\textbf{CBNA}$^{\myplus}$ & \cellcolor[gray]{.90} & \cellcolor[gray]{.90} & \cellcolor[gray]{.90} & \cellcolor[gray]{.90} & \cellcolor[gray]{.90} & \cellcolor[gray]{.90} & \cellcolor[gray]{.90} & \cellcolor[gray]{.90}\\
  %\hline
  \end{tabular}
  \label{tab:cross_dataset_comparison_syn}
\end{table*}

\begin{table}[t]
  \centering
  \setlength{\tabcolsep}{1.2pt}
  \caption{\textbf{Comparison to offline methods (UBNA \cite{Klingner2022})} across for \textbf{one real source domain and various real target domain datasets} showing the strong online capability of our CBNA algorithm. mIoU values in \%; best results written in \textbf{boldface}.}
  \begin{tabular}{c|l|ccc}
  %\hline
  & & & \\[-2.3ex]
   & \multirow{2}{*}{Method} & \multicolumn{3}{c}{\shortstack{$\mathcal{D}^{\mathrm{S}}$:\,Cityscapes; mIoU (19 classes)}} \\
  \cline{3-5}
  & & $\mathcal{D}^{\mathrm{T}}$:\,KITTI\stz & $\mathcal{D}^{\mathrm{T}}$:\,BDD\stz & $\mathcal{D}^{\mathrm{T}}$:\,Mapillary\stz \\
  \hline
  \multirow{5}{*}{\rotatebox{90}{\texttt{ResNet-50}\,}} & No adaptation\stz & 46.9 & \underline{36.6} & 43.0 \\
  \cline{2-5}
  & UBNA \cite{Klingner2022} (adapted to KITTI)\stz & \textbf{56.4} & 33.9 & 44.0 \\
  & UBNA \cite{Klingner2022} (adapted to BDD) & 47.6 & 35.9 & 40.4 \\
  & UBNA \cite{Klingner2022} (adapted to Mapillary) & 48.4 & 33.2 & \underline{44.3} \\
  \cline{2-5}
  & \cellcolor[gray]{.90} & \cellcolor[gray]{.90} & \cellcolor[gray]{.90}& \cellcolor[gray]{.90} \\[-2.3ex]
  %& \cellcolor[gray]{.90} \!\!\textbf{CBNA}$^{0.2}$ & 53.0 \cellcolor[gray]{.90} & 36.8 \cellcolor[gray]{.90} & 42.8 \cellcolor[gray]{.90} \\
  & \cellcolor[gray]{.90} \!\!\textbf{CBNA} & \underline{53.9}  \cellcolor[gray]{.90} & \textbf{38.3} \cellcolor[gray]{.90} & \textbf{45.8} \cellcolor[gray]{.90} \\
  %& \cellcolor[gray]{.90} \!\!\textbf{CBNA}$^{\myplus}$ & \cellcolor[gray]{.90} & \cellcolor[gray]{.90} & \cellcolor[gray]{.90} \\
  \hline
  \multirow{5}{*}{\rotatebox{90}{\texttt{VGG-16}\;}} & No adaptation\stz & 51.1 & \underline{32.7} & \underline{43.2} \\
  \cline{2-5}
  & UBNA \cite{Klingner2022} (adapted to KITTI)\stz & \textbf{57.1} & 28.5 & 37.8 \\ 
  & UBNA \cite{Klingner2022} (adapted to BDD) & 45.6 & 28.7 & 36.4 \\ 
  & UBNA \cite{Klingner2022} (adapted to Mapillary) & 46.7 & 28.5 & 36.1 \\
  \cline{2-5}
  & \cellcolor[gray]{.90} & \cellcolor[gray]{.90} & \cellcolor[gray]{.90} & \cellcolor[gray]{.90} \\[-2.3ex]
  %& \cellcolor[gray]{.90} \!\!\textbf{CBNA}$^{0.2}$ & 55.7 \cellcolor[gray]{.90} & 32.5 \cellcolor[gray]{.90} & 41.2 \cellcolor[gray]{.90} \\
   & \cellcolor[gray]{.90} \!\!\textbf{CBNA} & \underline{57.0} \cellcolor[gray]{.90} & \textbf{33.3} \cellcolor[gray]{.90} & \textbf{43.7} \cellcolor[gray]{.90} \\
  %& \cellcolor[gray]{.90} \!\!\textbf{CBNA}$^{\myplus}$ & \cellcolor[gray]{.90} & \cellcolor[gray]{.90} & \cellcolor[gray]{.90} \\
  \end{tabular}
  \label{tab:cross_dataset_comparison_real}
\end{table}
\par
In total, Table~\ref{tab:cross_dataset_comparison_syn} shows 16 adaptation conditions (2 source domains, 4 target domains, 2 network architectures). Here, our novel CBNA (one method!) secured in total twelve 1$^{\mathrm{st}}$ or 2$^{\mathrm{nd}}$ ranks, without any need of target domain data beforehand, while none of the four UBNA settings could achieve more than five such ranks. Presuming that the time of domain switch is always known, then all four methods ``UBNA adapted to X'' may be combined, achieving 16 1$^{\mathrm{st}}$ or 2$^{\mathrm{nd}}$ ranks, which performs comparable to our proposed CBNA, however, if the time of domain switch is not detected, then drastic performance decreases may occur. It is important to note that CBNA solves this issue with excellent computational efficiency, and does not suffer from adaptation mismatch. Accordingly, in only 3 out of 16 cases performance slightly decreased, while for UBNA there are many cases, where an adaptation mismatch leads to drastic decreases in performance. 
\par
To also answer the question, whether CBNA works when being applied to a model with significantly higher initial performance (real-to-real adaptation), we also experiment with models pre-trained on Cityscapes, as shown in Table~\ref{tab:cross_dataset_comparison_real}. Here, we again observe significant gains in performance, e.g., an absolute $7.0\%$, $1.7\%$, and $2.8\%$ on KITTI, BDD, and Mapillary, respectively, for the \texttt{ResNet-50} model. In comparison to the three UBNA methods, CBNA is always first or second ranked (6 such ranks) in any case better than the baseline without adaptation. The three UBNA methods \textit{together} only achieve 3 such ranks, often exhibiting an even decreased performance in the target domain.

\subsection{Method Performance Analysis}

While all presented results up to now can be applied without making use of source data, the question arises how CBNA, using only the source domain model and target data, compares to standard UDA methods, which make use of source and target domain data at once (no ``source-free adaptation'', not online capable). We provide such a comparison in Tables~\ref{tab:gta5_to_cityscapes} and \ref{tab:synthia_to_cityscapes} for the commonly used benchmarks GTA-5 to Cityscapes and SYNTHIA to Cityscapes, respectively.We compare to some of the latest state-of-the-art methods, where we observe that CBNA---as expected---performs worse than these UDA methods due to our much more constrained task definition. In a practical scenario, data often cannot be transferred from the model supplier, making CBNA the only method applicable to improve the model in such cases. Here, for a \texttt{VGG-16}-based model adaptation to Cityscapes ($\mathcal{D}^{\mathrm{T}}$), we observe improvements from $31.5\%$ to $36.4\%$ ($\mathcal{D}^{\mathrm{S}}$: GTA-5) and from $30.0\%$ to $32.1\%$/from $35.2\%$ to $37.7\%$ ($\mathcal{D}^{\mathrm{S}}$: SYNTHIA). Similar improvements are achieved with a \texttt{ResNet-50} backbone.

% %%%%%%%%%%%%%Synthetic-to-Real%%%%%%%%
% %%%%%%%%%%%%%%%%%%%%%
% % To Cityscapes

\begin{figure}[t!]
 	\centering	
 	%\resizebox{\linewidth}{!}{\input{figs/method_analysis/performance_histogramm.pgf}}
 	\includegraphics[width=\linewidth]{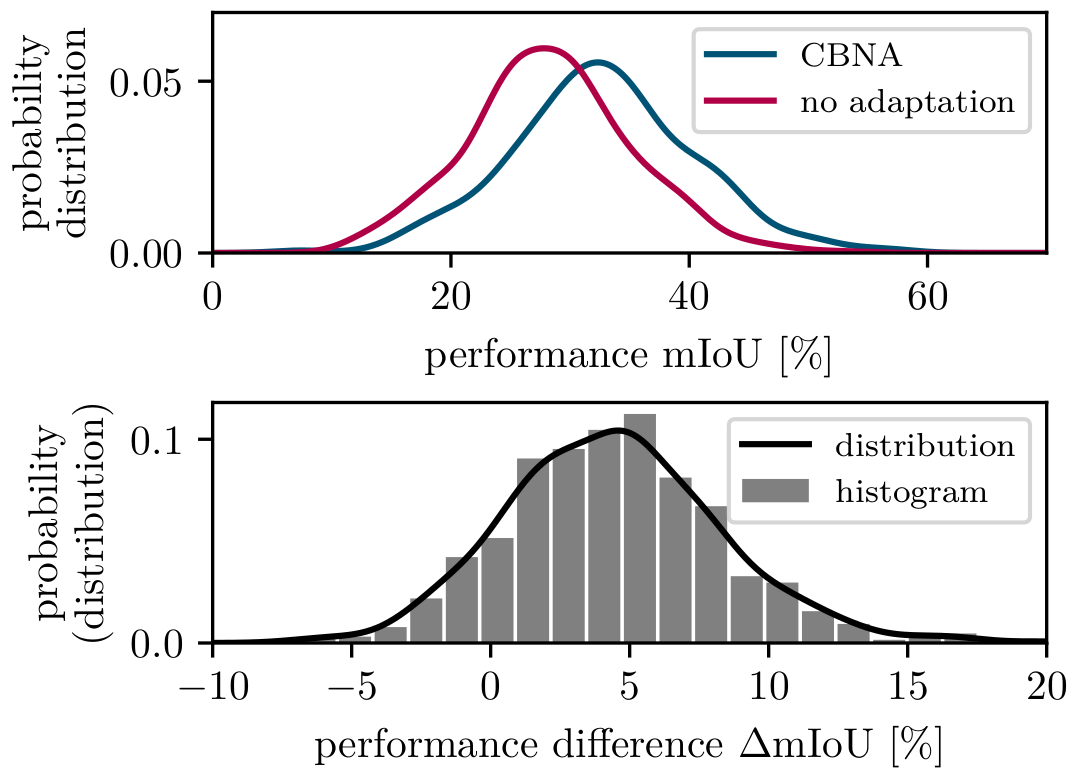}
 	\caption{\revision{1.6}{Probability distributions over the absolute single-image performances (top) and the single-image performance difference before and after application of CBNA (bottom) when adapting from GTA-5 ($\mathcal{D}^{\mathrm{S}}$) to Cityscapes ($\mathcal{D}^{\mathrm{T}}$) using a \texttt{VGG-16} backbone.}}
 	\label{fig:miou_hist}
 \end{figure}

\begin{table*}[t]
  \centering
  \caption{\textbf{Comparison to UDA methods} on the \textbf{Cityscapes validation set} for the adaptation from \textbf{GTA-5} ($\mathcal{D}^{\mathrm{S}}$) \textbf{to Cityscapes} ($\mathcal{D}^{\mathrm{T}}$). Best UDA results and best source-free UDA results in boldface; results marked with $*$ are taken from the respective publications.}
  \vspace{-0.17cm}
  \footnotesize
  \setlength{\tabcolsep}{1.35pt}
  \begin{tabular}{c|l|c|c|ccccccccccccccccccc|c}
   \rotatebox{90}{Network} & Method  & \shortstack{Source-\\ free \\ adaptation} & \shortstack{Online\\ capable} & \rotatebox{90}{road} & \rotatebox{90}{sidewalk} & \rotatebox{90}{building} & \rotatebox{90}{wall} & \rotatebox{90}{fence} & \rotatebox{90}{pole} & \rotatebox{90}{traffic light} & \rotatebox{90}{traffic sign} & \rotatebox{90}{vegetation} & \rotatebox{90}{terrain} & \rotatebox{90}{sky} & \rotatebox{90}{person} & \rotatebox{90}{rider} & \rotatebox{90}{car} & \rotatebox{90}{truck} & \rotatebox{90}{bus} & \rotatebox{90}{on rails} & \rotatebox{90}{motorbike} & \rotatebox{90}{bike} & \shortstack{$\mathrm{mIoU}$ \\ (19 cl.)} \\
  \hline
  \hline
  \multirow{11}{*}{\rotatebox{90}{\texttt{ResNet-50}}}  & Dong~et~al.~\cite{Dong2020}$^*$\stz & no & no &
  89.6 & 50.4 & 83.0 & 35.6 & 26.9 & 31.1 & 37.3 & 35.1 & 83.5 & \textbf{40.6} & 84.0 & 60.6 & \textbf{34.3} & 80.9 & 35.1 & 47.3 & 0.5 & 34.5 & 33.7 & 48.6 \\
   & Kim~et~al.~\cite{Kim2020}$^*$ & no & no &  
  92.9 & 55.0 &  85.3 & 34.2 & \textbf{31.1} & \textbf{34.9} & 40.7 & 34.0 & \textbf{85.2} & 40.1 & 87.1 & 61.0 & 31.1 & 82.5 & 32.3 & 42.9 & 0.3 & \textbf{36.4} & \textbf{46.1} & 50.2 \\
   & Mei~et~al.~\cite{Mei2020}$^*$ & no & no & 
  \textbf{94.1} & \textbf{58.8} & \textbf{85.4} & \textbf{39.7} & 29.2 & 25.1 & \textbf{43.1} & \textbf{34.2} & 84.8 & 34.6 & \textbf{88.7} & \textbf{62.7} & 30.3 & \textbf{87.6} & \textbf{42.3} & \textbf{50.3} & \textbf{24.7} & 35.2 & 40.2 & \textbf{52.2} \\
  \cline{2-24}
   & No adaptation\stz  & - & - & 58.1 & 23.8 & 70.5 & 14.8 & 19.2 & 30.5 & 29.0 & 17.7 & 79.1 & 21.8 & 83.1 & 56.4 & 14.8 & 72.3 & 19.5 & 4.5 & 0.9 & 16.5 & 5.6 & 33.6 \\
   & \revision{1.1}{Termöhlen~et~al.~\cite{Termoehlen2021}$^*$} & no & yes & 83.5 & 28.0 & 75.9 & 18.0 & 22.2 & 30.3 & 30.6 & 19.4 & 82.0 & 34.4 & 71.7 & 56.3 & \textbf{25.3} & 71.4 & 21.7 & 33.5 & 0.1 & 28.2 & 33.0 & 40.3 \\
   & Klingner~et~al.~\cite{Klingner2022}$^*$ & yes & no & 81.8 & 32.3 & 79.5 & 18.2 & 23.8 & 34.9 & 29.5 & 19.8 & 74.2 & 17.9 & 82.4 & 57.5 & 11.1 & 81.6 & 16.1 & 19.0 & 2.5 & 21.3 & 9.8 & 37.5\\
   & \revision{1.1}{Liu~et~al.~\cite{Liu2021}$^*$} & yes & no & 84.2 & 39.2 & 82.7 & 27.5 & 22.1 & 25.9 & 31.1 & 21.9 & 82.4 & 30.5 & 85.3 & 58.7 & 22.1 & 80.0 & \textbf{33.1} & 31.5 & 3.6 & \textbf{27.8} & 30.6 & 43.2\\
   & \revision{1.1}{Teja~et~al.~\cite{Teja2021}$^*$} & yes & no & \textbf{92.3} & \textbf{55.2} & 81.6 & \textbf{30.8} & 18.8 & 37.1 & 17.7 & 12.1 & 84.2 & \textbf{35.9} & 83.8 & 57.7 & 24.1 & 81.7 & 27.5 & \textbf{44.3} & 6.9 & 24.1 & \textbf{40.4} & \textbf{45.1}\\
   & \cellcolor[gray]{.90}\textbf{CBNA}\stz& \cellcolor[gray]{.90}yes& \cellcolor[gray]{.90}yes& \cellcolor[gray]{.90}69.9& \cellcolor[gray]{.90}25.8& \cellcolor[gray]{.90}78.8& \cellcolor[gray]{.90}20.9& \cellcolor[gray]{.90}23.5& \cellcolor[gray]{.90}34.1& \cellcolor[gray]{.90}27.5& \cellcolor[gray]{.90}16.0& \cellcolor[gray]{.90}79.6& \cellcolor[gray]{.90}23.5& \cellcolor[gray]{.90}82.8& \cellcolor[gray]{.90}56.7& \cellcolor[gray]{.90}12.6& \cellcolor[gray]{.90}81.3& \cellcolor[gray]{.90}20.9& \cellcolor[gray]{.90}16.6& \cellcolor[gray]{.90}0.5& \cellcolor[gray]{.90}17.5& \cellcolor[gray]{.90}8.2& \cellcolor[gray]{.90}36.7 \\
   & No adaptation (DG-Init)\stz  & - & - & 75.9 & 40.7 & 74.4 & 12.0 & 20.9 & 35.6 & 31.1 & 42.5 & 84.0 & 17.3 & 87.4 & 56.9 & 14.7 & 78.4 & 24.3 & 2.8 & 0.1 & 11.4 & 18.1 & 38.3 \\
   & \cellcolor[gray]{.90}\textbf{CBNA} (DG-Init)& \cellcolor[gray]{.90}yes& \cellcolor[gray]{.90}yes& \cellcolor[gray]{.90}89.4 & \cellcolor[gray]{.90}48.4 & \cellcolor[gray]{.90}\textbf{83.8}& \cellcolor[gray]{.90}21.1 & \cellcolor[gray]{.90}\textbf{26.1}& \cellcolor[gray]{.90}\textbf{42.8}& \cellcolor[gray]{.90}\textbf{35.5}& \cellcolor[gray]{.90}\textbf{45.0}& \cellcolor[gray]{.90}\textbf{85.3}& \cellcolor[gray]{.90}32.2 & \cellcolor[gray]{.90}\textbf{88.9}& \cellcolor[gray]{.90}\textbf{60.1}& \cellcolor[gray]{.90}21.3 & \cellcolor[gray]{.90}\textbf{85.9}& \cellcolor[gray]{.90}25.3 & \cellcolor[gray]{.90}6.2& \cellcolor[gray]{.90}\textbf{10.9}& \cellcolor[gray]{.90}14.6& \cellcolor[gray]{.90}27.7 & \cellcolor[gray]{.90}44.8 \\
  \hline
  \hline
  \multirow{9}{*}{\rotatebox{90}{\texttt{VGG-16}}} & Dong~et~al.~\cite{Dong2020}$^*$\stz & no & no & 
  89.8 & 46.1 & 75.2 & 30.1 & \textbf{27.9} & 15.0 & 20.4 & 18.9 & 82.6 & 39.1 & 77.6 & 47.8 & 17.4 & 76.2 & 28.5 & \textbf{33.4} & 0.5 & 29.4 & \textbf{30.8} & 41.4 \\
   & Kim~et~al.~\cite{Kim2020}$^*$ & no & no & 
  \textbf{92.5} & \textbf{54.5} & \textbf{83.9} & \textbf{34.5} & 25.5 & \textbf{31.0} & 30.4 & 18.0 & \textbf{84.1} & \textbf{39.6} & \textbf{83.9} & 53.6 & 19.3 & 81.7 & 21.1 & 13.6 & \textbf{17.7} & 12.3 & 6.5 & 42.3 \\
   & Yang~et~al.~\cite{Yang2020a}$^*$ & no & no & 
  90.1 & 41.2 & 82.2 & 30.3 & 21.3 & 18.3 & \textbf{33.5} & \textbf{23.0} & \textbf{84.1} & 37.5 & 81.4 & \textbf{54.2} & \textbf{24.3} & \textbf{83.0} & \textbf{27.6} & 32.0 & 8.1 & \textbf{29.7} & 26.9 & \textbf{43.6} \\
  \cline{2-24}
   & No adaptation\stz & - & - & 55.8 & 21.9 & 65.9 & 15.2 & 14.7 & 27.5 & 31.0 & 17.9 & 77.8 & 19.5 & 74.4 & 55.2 & 12.1 & 71.7 & 11.9 & 3.3 & 0.5 & 13.2 & 9.6 & 31.5\\
   & Klingner~et~al.~\cite{Klingner2022}$^*$ & yes & no & 80.8 & 29.4 & 77.6 & 19.8 & 17.1 & 33.9 & 29.3 & 20.5 & 73.9 & 16.8 & 76.7 & 58.3 & \textbf{15.2} & 79.1 & 13.6 & 12.5 & 5.7 & \textbf{14.1} & 10.8 & 36.1\\
   & \revision{1.1}{Liu~et~al.~\cite{Liu2021}$^*$} & yes & no & \textbf{81.8} & 35.4 & \textbf{82.3} & \textbf{21.6} & 20.2 & 25.3 & 17.8 & 4.7 & 80.7 & 24.6 & 80.4 & 50.5 & 9.2 & 78.4 & \textbf{26.3} & \textbf{19.8} & \textbf{11.1} & 6.7 & 4.3 & 35.9\\
   & \cellcolor[gray]{.90}\textbf{CBNA}\stz& \cellcolor[gray]{.90}yes& \cellcolor[gray]{.90}yes& \cellcolor[gray]{.90}75.8& \cellcolor[gray]{.90}31.9& \cellcolor[gray]{.90}75.5& \cellcolor[gray]{.90}17.2& \cellcolor[gray]{.90}17.9& \cellcolor[gray]{.90}34.4& \cellcolor[gray]{.90}30.0& \cellcolor[gray]{.90}18.9& \cellcolor[gray]{.90}80.5& \cellcolor[gray]{.90}22.7& \cellcolor[gray]{.90}78.0& \cellcolor[gray]{.90}58.3& \cellcolor[gray]{.90}14.0& \cellcolor[gray]{.90}\textbf{82.6}& \cellcolor[gray]{.90}15.2& \cellcolor[gray]{.90}10.4& \cellcolor[gray]{.90}1.5& \cellcolor[gray]{.90}13.2& \cellcolor[gray]{.90}13.0& \cellcolor[gray]{.90}36.4 \\
   & No adaptation (DG-Init)\stz  & - & - & 64.7 & 32.8 & 73.6 & 16.5 & 22.8 & 39.4 & 37.0 & 44.6 & 85.9 & 30.8 & 83.9 & 58.1 & 07.2 & 68.3 & 18.3 & 8.1 & 5.2 & 9.7 & 13.8 & 37.9 \\
   & \cellcolor[gray]{.90}\textbf{CBNA} (DG-Init)& \cellcolor[gray]{.90}yes& \cellcolor[gray]{.90}yes& \cellcolor[gray]{.90}81.4 & \cellcolor[gray]{.90}\textbf{41.5}& \cellcolor[gray]{.90}81.8 & \cellcolor[gray]{.90}21.1 & \cellcolor[gray]{.90}\textbf{26.0}& \cellcolor[gray]{.90}\textbf{44.2}& \cellcolor[gray]{.90}\textbf{41.3}& \cellcolor[gray]{.90}\textbf{45.0}& \cellcolor[gray]{.90}\textbf{86.5}& \cellcolor[gray]{.90}\textbf{35.0}& \cellcolor[gray]{.90}\textbf{87.1}& \cellcolor[gray]{.90}\textbf{60.6}& \cellcolor[gray]{.90}14.8& \cellcolor[gray]{.90}80.7& \cellcolor[gray]{.90}22.9 & \cellcolor[gray]{.90}12.4& \cellcolor[gray]{.90}5.8 & \cellcolor[gray]{.90}12.6 &\cellcolor[gray]{.90}\textbf{19.3}& \cellcolor[gray]{.90}\textbf{43.1}\\
  \end{tabular}
  \label{tab:gta5_to_cityscapes}
\end{table*}

\begin{table*}[t]
  \centering
  \caption{\textbf{Comparison to UDA methods} on the \textbf{Cityscapes validation set} for the adaptation from \textbf{SYNTHIA} ($\mathcal{D}^{\mathrm{S}}$) \textbf{to Cityscapes} ($\mathcal{D}^{\mathrm{T}}$). Best UDA results and best source-free UDA results in boldface; results marked with $*$ are taken from the respective publications.}
  \vspace{-0.17cm}
  \footnotesize
  \setlength{\tabcolsep}{1.0pt}
  \begin{tabular}{c|l|c|c|ccccccccccccccccccc|c|c}
   \rotatebox{90}{Network} & Method & \shortstack{Source-\\ free \\ adaptation} & \shortstack{Online\\ capable} & \rotatebox{90}{road} & \rotatebox{90}{sidewalk} & \rotatebox{90}{building} & \rotatebox{90}{wall} & \rotatebox{90}{fence} & \rotatebox{90}{pole} & \rotatebox{90}{traffic light} & \rotatebox{90}{traffic sign} & \rotatebox{90}{vegetation} &  \rotatebox{90}{terrain} & \rotatebox{90}{sky} & \rotatebox{90}{person} & \rotatebox{90}{rider} & \rotatebox{90}{car} & \rotatebox{90}{truck} & \rotatebox{90}{bus} & \rotatebox{90}{on rails} & \rotatebox{90}{motorbike} & \rotatebox{90}{bike} & \shortstack{$\mathrm{mIoU}$ \\ (16 cl.)} & \shortstack{$\mathrm{mIoU}$ \\ (13 cl.)}\\
  \hline
  \hline
  \multirow{11}{*}{\rotatebox{90}{\texttt{ResNet-50}}} & Yang~et~al.~\cite{Yang2020a}$^*$\stz & no & no &
  85.1 & 44.5 & 81.0 & - & - & - & 16.4 & 15.2 & 80.1 & - & 84.8 & 59.4 & 31.9 & 73.2 & - & 41.0 & - & 32.6 & 44.7 & - & 53.1 \\
   & Dong~et~al.~\cite{Dong2020}$^*$ & no & no &
  80.2 & 41.1 & 78.9 & \textbf{23.6} & 0.6 & 31.0 & 27.1 & \textbf{29.5} & 82.5 & - & 83.2 & 62.1 & 26.8 & 81.5 & - & 37.2 & - & 27.3 & 42.9 & 47.2 & - \\
   & Mei~et~al.~\cite{Mei2020}$^*$ & no & no &
  81.9 & 41.5 & \textbf{83.3} & 17.7 & \textbf{4.6} & \textbf{32.3} & \textbf{30.9} & 28.8 & \textbf{83.4} & - & 85.0 & \textbf{65.5} & 30.8 & \textbf{86.5} & - & 38.2 & - & \textbf{33.1} & \textbf{52.7} & \textbf{49.8} & \textbf{57.0} \\
  \cline{2-25}
   & No adaptation\stz & - & - & 36.5 & 18.6 & 68.3 & 2.0 & 0.2 & 30.3 & 6.0 & 10.2 & 74.5 &  - & 81.6 & 51.9 & 10.6 & 41.3 & - & 9.5 & - & 2.2 & 22.6 & 29.1 & 34.1 \\
   & \revision{1.1}{Termöhlen~et~al.~\cite{Termoehlen2021}$^*$} & no & yes & 63.6 & 24.0 & 65.7 & - & - & - & 4.3 & 13.7 & 62.5 & - & 77.2 & 54.8 & \textbf{20.0} & 62.1 & - & 9.3 & - & 15.5 & 29.9 & - & 38.7 \\
   & Klingner~et~al.~\cite{Klingner2022}$^*$ & yes & no & 62.5 & 22.8 & 75.6 & 3.1 & 0.5 & 32.5 & 8.6 & 11.3 & 73.0 & - & 82.7 & 42.5 & 12.5 & 67.1 &  -  & 12.5 & - & 5.7 & 27.8 & 33.8 & 39.7 \\
   & \revision{1.1}{Teja~et~al.~\cite{Teja2021}$^*$} & yes & no & 59.3 & 24.6 & 77.0 & \textbf{14.0} & 1.8 & 31.5 & 18.3 & 32.0 & 83.1 & - & 80.4 & 46.3 & 17.8 & 76.7 & - & 17.0 & - & \textbf{18.5} & 34.6 & 39.6 & 45.0\\
   & \revision{1.1}{Liu~et~al.~\cite{Liu2021}$^*$} & yes & no & \textbf{81.9} & 44.9 & \textbf{81.7} & 4.0 & 0.5 & 26.2 & 3.3 & 10.7 & \textbf{86.3} & - & 89.4 & 37.9 & 13.4 & \textbf{80.6} & - & \textbf{25.6} & - & 9.6 & 31.3 & 39.2 & 45.9\\
   & \cellcolor[gray]{.90}\textbf{CBNA}\stz& \cellcolor[gray]{.90}yes& \cellcolor[gray]{.90}yes& \cellcolor[gray]{.90}53.9& \cellcolor[gray]{.90}21.6& \cellcolor[gray]{.90}74.5& \cellcolor[gray]{.90}1.2& \cellcolor[gray]{.90}0.2& \cellcolor[gray]{.90}33.4& \cellcolor[gray]{.90}7.9& \cellcolor[gray]{.90}12.4& \cellcolor[gray]{.90}77.4& \cellcolor[gray]{.90}-& \cellcolor[gray]{.90}81.5& \cellcolor[gray]{.90}42.7& \cellcolor[gray]{.90}11.7& \cellcolor[gray]{.90}57.0& \cellcolor[gray]{.90}-& \cellcolor[gray]{.90}12.2& \cellcolor[gray]{.90}-& \cellcolor[gray]{.90}4.9& \cellcolor[gray]{.90}27.6& \cellcolor[gray]{.90}32.5& \cellcolor[gray]{.90}38.2 \\
   & No adaptation (DG-Init)\stz  & - & - & 70.6 & 41.0 & 71.9 & 10.2 & \textbf{14.6} & 40.6 & 26.4 & \textbf{40.0} & 85.6 & - & \textbf{90.4} & \textbf{60.0} & 17.5 & 54.0 & - & 10.3 & - & 4.2& 36.3& 42.1 & 47.7 \\
   & \cellcolor[gray]{.90}\textbf{CBNA} (DG-Init)& \cellcolor[gray]{.90}yes& \cellcolor[gray]{.90}yes& \cellcolor[gray]{.90}79.9 & \cellcolor[gray]{.90}\textbf{46.7}& \cellcolor[gray]{.90}74.5& \cellcolor[gray]{.90}10.5 & \cellcolor[gray]{.90}10.2& \cellcolor[gray]{.90}\textbf{41.3}& \cellcolor[gray]{.90}\textbf{28.3}& \cellcolor[gray]{.90}39.1& \cellcolor[gray]{.90}84.3& \cellcolor[gray]{.90}-& \cellcolor[gray]{.90}88.6& \cellcolor[gray]{.90}59.0& \cellcolor[gray]{.90}18.5 & \cellcolor[gray]{.90}75.8 & \cellcolor[gray]{.90}-& \cellcolor[gray]{.90}14.5 & \cellcolor[gray]{.90}-& \cellcolor[gray]{.90}4.1& \cellcolor[gray]{.90}\textbf{37.0}& \cellcolor[gray]{.90}\textbf{44.5}& \cellcolor[gray]{.90}\textbf{51.1} \\
  \hline
  \hline
  \multirow{8}{*}{\rotatebox{90}{\texttt{VGG-16}}}  & Lee~et~al.~\cite{Lee2019d}$^*$\stz & no & no & 
  71.1 & 29.8 & 71.4 & 3.7 & 0.3 & \textbf{33.2} & 6.4 & 15.6 & \textbf{81.2} & - & 78.9 & 52.7 & 13.1 & 75.9 & - & 25.5 & - & 10.0 & 20.5 & 36.8 & \textbf{42.4}\\
   & Dong~et~al.~\cite{Dong2020}$^*$ & no & no &  
  70.9 & \textbf{30.5} & \textbf{77.8} & \textbf{9.0} & \textbf{0.6} & 27.3 & 8.8 & 12.9 & 74.8 & - & 81.1 & 43.0 & \textbf{25.1} & 73.4 & - & \textbf{34.5} & - & \textbf{19.5} & 38.2 & 39.2 & - \\
   & Yang~et~al.~\cite{Yang2020a}$^*$ & no & no & 
  \textbf{73.7} & 29.6 & 77.6 & 1.0 & 0.4 & 26.0 & \textbf{14.7} & \textbf{26.6} & 80.6 & - & \textbf{81.8} & \textbf{57.2} & 24.5 & \textbf{76.1} & - & 27.6 & - & 13.6 & \textbf{46.6} & \textbf{41.1} & - \\
  \cline{2-25}
   & No adaptation\stz & - & - & 49.4 & 20.8 & 61.5 & 3.6 & 0.1 & 30.5 & 13.6 & 14.1 & 74.4 & - & 75.5 & 53.5 & 10.6 & 47.2 & - & 4.8 & - & 3.0 & 17.1 & 30.0 & 35.2 \\
   & Klingner~et~al.~\cite{Klingner2022}$^*$ & yes & no & 72.3 & 26.6 & 73.0 & 2.3 & 0.3 & 31.5 & 12.1 & 16.6 & 72.1 & - & 75.6 & 45.4 & 13.6 & 61.2 & - & \textbf{8.5} & - & \textbf{8.5} & 30.1 & 34.4 & 40.7 \\
   %\cline{2-25}
   & \cellcolor[gray]{.90}\textbf{CBNA}\stz& \cellcolor[gray]{.90}yes& \cellcolor[gray]{.90}yes& \cellcolor[gray]{.90}52.9& \cellcolor[gray]{.90}25.0& \cellcolor[gray]{.90}62.4& \cellcolor[gray]{.90}2.7& \cellcolor[gray]{.90}0.2& \cellcolor[gray]{.90}32.5& \cellcolor[gray]{.90}13.3& \cellcolor[gray]{.90}16.1& \cellcolor[gray]{.90}78.8& \cellcolor[gray]{.90}-& \cellcolor[gray]{.90}75.8& \cellcolor[gray]{.90}46.8& \cellcolor[gray]{.90}12.8& \cellcolor[gray]{.90}55.7& \cellcolor[gray]{.90}-& \cellcolor[gray]{.90}6.3& \cellcolor[gray]{.90}-& \cellcolor[gray]{.90}6.9& \cellcolor[gray]{.90}26.0& \cellcolor[gray]{.90}32.1& \cellcolor[gray]{.90}37.7 \\
   & No adaptation (DG-Init)\stz  & - & - & 72.3 & 45.3 & \textbf{78.9} & \textbf{12.6} & \textbf{15.0} & 41.1 & 28.4 & \textbf{42.7} & \textbf{85.3} & - & 87.8 & 61.5 & 20.0 & 50.5 & - & 4.5 & - & 4.0 & 40.2 & 43.1 & 48.7 \\
   & \cellcolor[gray]{.90}\textbf{CBNA} (DG-Init)& \cellcolor[gray]{.90}yes& \cellcolor[gray]{.90}yes& \cellcolor[gray]{.90}\textbf{77.9}& \cellcolor[gray]{.90}\textbf{48.1}& \cellcolor[gray]{.90}78.8& \cellcolor[gray]{.90}11.2& \cellcolor[gray]{.90}8.0& \cellcolor[gray]{.90}\textbf{43.6}& \cellcolor[gray]{.90}\textbf{31.7}& \cellcolor[gray]{.90}42.2& \cellcolor[gray]{.90}84.3& \cellcolor[gray]{.90}-& \cellcolor[gray]{.90}\textbf{88.2}& \cellcolor[gray]{.90}\textbf{62.7}& \cellcolor[gray]{.90}\textbf{21.3}& \cellcolor[gray]{.90}\textbf{63.2}& \cellcolor[gray]{.90}-& \cellcolor[gray]{.90}5.5& \cellcolor[gray]{.90}-& \cellcolor[gray]{.90}7.1& \cellcolor[gray]{.90}\textbf{41.6}& \cellcolor[gray]{.90}\textbf{44.7}& \cellcolor[gray]{.90}\textbf{51.3} \\
  \end{tabular}
  \label{tab:synthia_to_cityscapes}
\end{table*}

\par
A particular advantage is that CBNA can be combined with any domain generalization (DG) pre-training, which is not necessarily the case for standard UDA methods. This would allow a supplier to improve the model, while the car manufacturer can still improve the final model performance on the target domain during vehicle operation using CBNA. Exploiting this advantage, we also present results for such a DG-initialized model, where for \texttt{VGG-16}-based models and the adaptation from GTA-5 or SYNTHIA to Cityscapes, we achieve significantly higher performances of $43.1\%$ and $44.7\%$/$51.3\%$, respectively, than without DG initialization ($36.4\%$ and $32.1\%$/$37.7\%$). This reduces the gap to UDA methods, which are sometimes even outperformed by the combination of DG pre-training and CBNA as, e.g., for a \texttt{VGG-16}-based model and the adaptation from SYNTHIA to Cityscapes. In other cases, the final performance of CBNA with DG pre-training is still slightly worse than UDA methods, but offers a good alternative to UDA methods, when simultaneous access to source and target domain data is not possible.
\par
\revision{1.6}{As our method is applicable on a single-image basis, we further analyze the single-image performance in Fig.~\ref{fig:miou_hist}. In the top part, we plot the distributions over the absolute performance for different models. We observe a clear shift towards a higher performance for CBNA compared to the no adaptation model. We further compare the performance before and after application of CBNA for single images and plot the distribution over this performance difference in the bottom part of Fig.~\ref{fig:miou_hist}. We can see that for the large majority of images CBNA improves performance, but for some images the performance also decreases slightly.}
\par
\begin{figure}[t]
	\centering	
	\includestandalone[width=1.0\linewidth]{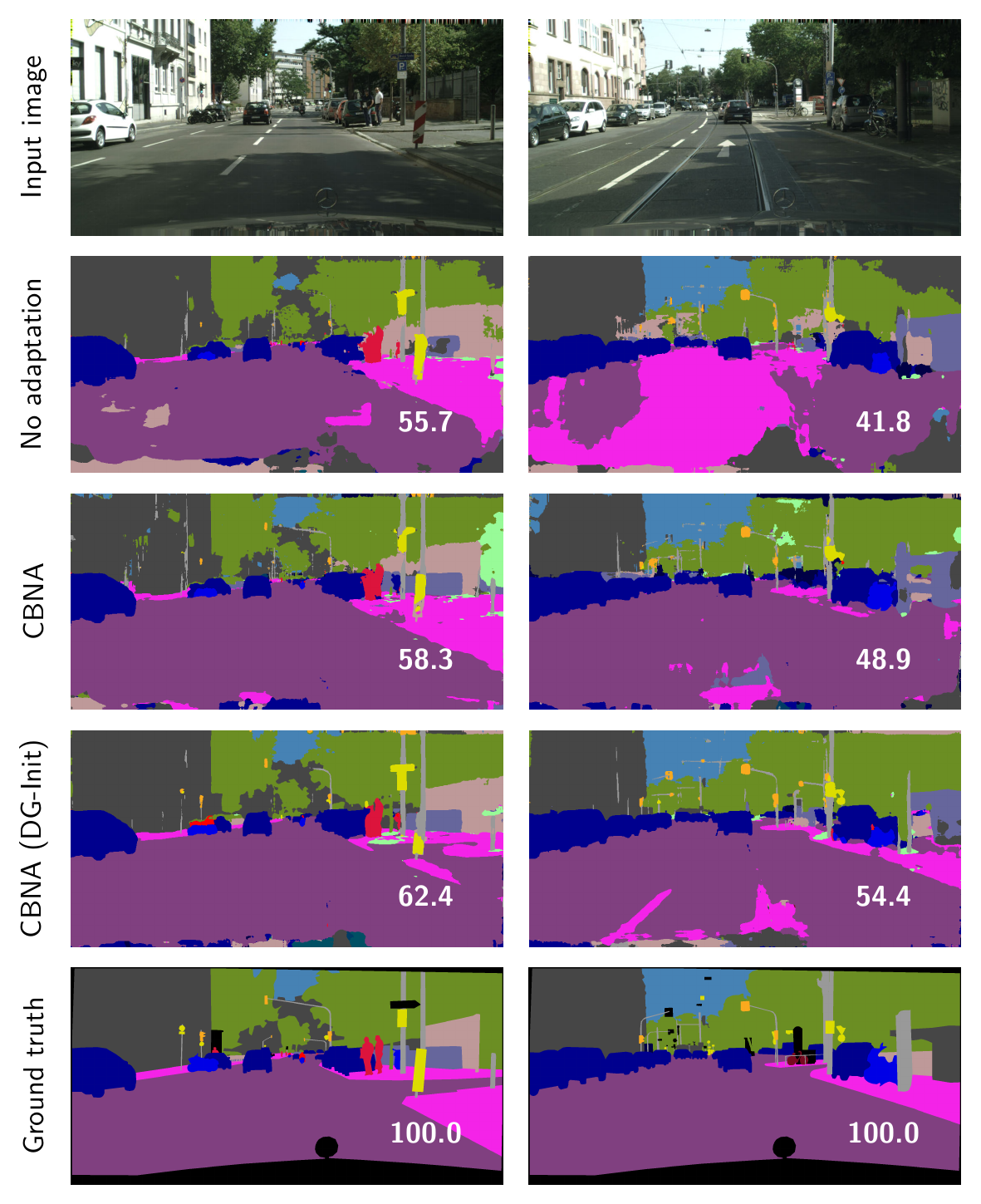}
	\caption{Qualitative comparison when adapting from GTA-5 ($\mathcal{D}^{\mathrm{S}}$) to Cityscapes ($\mathcal{D}^{\mathrm{T}}$) between the model without adaptation and our CBNA models using a \texttt{VGG-16} backbone. Single image mIoU performance [\%] in white.}
	\label{fig:qualitative}
\end{figure}

The mentioned improvements are also illustrated in Fig.~\ref{fig:qualitative}, where the segmentation masks generated by CBNA contain much fewer artifacts than the ones of the ``no adaptation'' baseline. Using a model that has been pre-trained using a DG pre-training improves the results even further, which is consistent with the results from Tables~\ref{tab:gta5_to_cityscapes} and \ref{tab:synthia_to_cityscapes}. 

\section{Conclusions}
\label{sec:conclusion}

We presented a continual domain adaptation method for semantic segmentation in constrained (practical) scenarios, where one does not have simultaneous access to both source and target domain data. For these cases, the given trained deep neural network (DNN) model from the source domain can be adapted in an online fashion to single images of different target domains by using our source-free Continuous BatchNorm Adaptation (CBNA) method, which yields a significant increase in performance. This presents a clear advantage over previous offline unsupervised domain adaptation (UDA) methods, as we perform a single-image adaptation which is employed during inference, requiring only minimal computational overhead while incurring no algorithmic delay. For semantic segmentation, we presented experiments for three source domains and four target domains, showing the good generalization capability of our method. We thereby offer the possibility to deploy a UDA method in an online fashion (i.e., in an operating vehicle) for continual adaptation. Future works could integrate our method in tasks such as instance segmentation or object detection as a standard normalization layer modification to improve these tasks' target domain performance. Also, transferring the advances of other source-free domain adaptation methods to the continual setting may further facilitate target domain performance gains.

\appendix

\subsection{Mapillary Label Inconsistency}
\label{sec:label_inconsistency}

\revision{1.5}{To deal with the label inconsistency between Cityscapes and Mapillary, the classes “bike-lane”, “crosswalk-plain”, “road”, “lane marking - crosswalk”, and “lane marking - general” are mapped to the “road” class, and the classes “bicyclist”, “motorcyclist”, and “other rider” are mapped to the “rider” class. All other classes defined in Cityscapes are also present in Mapillary and can be mapped in a straightforward fashion. All remaining additional classes defined in Mapillary are mapped to the background class.}

\subsection{Network Architecture Details}
\label{sec:network_architecture}

\revision{1.4}{We rely on the encoder-decoder network architecture from \cite{Klingner2022}. The encoder is a standard \texttt{ResNet-50} \cite{He2016} or \texttt{VGG-16} \cite{Simonyan2015} model with Imagenet-pretrained weights \cite{Russakovsky2015}. The basic setup of each layer in these architectures is the use of a convolutional layer, followed by a batch normalization (BN) layer, followed by an activation function (mainly ReLU variants), where the BN layers in the encoder are essential for our adaptation method. In total, the feature resolution is downsampled five times resulting in a downsampling factor of $2^5$. The intermediate features at each resolution are passed on to the decoder, implementing a U-Net-like structure inspired by \cite{Ronneberger2015}.} 
\par
\revision{1.4}{The decoder uses a simple fully convolutional architecture as defined in \cite{Godard2019}. The features from the skip connections are concatenated with the features from the decoder, followed by two convolutional layers with ELU activation and nearest neighbor upsampling. Note that no BN layers are used in the decoder. The output convolution produces output logits in $S= |\mathcal{S}|$ feature maps, which are converted to posterior class probabilities for each pixel by a pixel-wise softmax function.}

\subsection{Method Complexity Analysis}

\begin{table}[t]
  \centering
  \setlength{\tabcolsep}{1pt}
  \caption{\textbf{Additional FLOPs} induced by the single equations involved for CBNA and for the reference methods C-X.}
  \begin{tabular}{c|l|c|c|c|c|c}
  \multicolumn{2}{l|}{Equations} & (\ref{eq:mean}), (\ref{eq:variance}) & (\ref{eq:ubna_mean}), (\ref{eq:ubna_variance}) & (\ref{eq:cbna_feature_mean}), (\ref{eq:cbna_feature_variance}) & (\ref{eq:cbna_mean}), (\ref{eq:cbna_variance}) & forward pass\\
  \hline
  \multirow{3}{*}{\rotatebox{90}{\!\!FLOPs}} & Complexity & $\sim \!\!H_{\ell} W_{\ell} C_{\ell}$\stz & $\sim \!C_{\ell}$\stz & $\sim \!\!H_{\ell} W_{\ell} C_{\ell}$\stz & $\sim \!C_{\ell}$\stz & \\
  \cline{2-7}
   & \texttt{ResNet-50} & $300\cdot 10^6$\stz & $17\cdot 10^3$\stz & $300\cdot 10^6$\stz & $17\cdot 10^3$\stz & $43\cdot10^9$\stz \\
   & \texttt{VGG-16}  & $430\cdot 10^6$ & $106\cdot 10^3$ & $430\cdot 10^6$ & $106\cdot 10^3$ & $161\cdot10^9$ \\
   \hline
   \multicolumn{2}{l|}{Methods} & C-Klingner & C-Klingner & \textbf{CBNA} & \textbf{CBNA} & C-Klingner\\
   \multicolumn{2}{c|}{}& C-Li & &  &  & \\
   \multicolumn{2}{c|}{}& C-Zhang & & & & \\
  \end{tabular}
  \label{tab:complexity_analysis_app}
\end{table}

To better understand the additional computational complexity induced by CBNA and the reference methods C-X, we analyze the single involved equations in terms of their induced additional FLOPs in Table~\ref{tab:complexity_analysis_app}. The numbers are accumulated over all BN layers in the \texttt{VGG-16} and \texttt{ResNet-50} network architectures. For the reference methods C-Li and C-Zhang, we need to apply (\ref{eq:mean}) and (\ref{eq:variance}), which then replace the source domain statistics. As here the mean over each feature map is computed, the additional FLOPs induced by these equations scale with the feature map resolution $H_{\ell}\cdot W_{\ell}$ and the number of feature maps $C_{\ell}$. 
\par
For C-Klingner, additionally (\ref{eq:ubna_mean}) and (\ref{eq:ubna_variance}) are applied on the first forward pass to mix source and target domain statistics. As, however, only one value per feature map is updated, these equations induce additional FLOPs in the order of $C_{\ell}$. However, here most additional computations are induced by the additional forward pass (cf.~Table~\ref{tab:complexity_analysis_app}), where all computations in the convolutional layers have to be recomputed, inducing much more additional FLOPs than just a few additional computations in the BN layers as in C-Li/C-Zhang.
\par 
For CBNA, on the other hand, first (\ref{eq:cbna_feature_mean}) and (\ref{eq:cbna_feature_variance}) have to be applied, which, however, induces exactly the same number of additional FLOPs as (\ref{eq:mean}) and (\ref{eq:variance}) for C-Li/C-Zhang, i.e., (\ref{eq:cbna_feature_mean}) and (\ref{eq:cbna_feature_variance}) also scale with $H_{\ell}\cdot W_{\ell}\cdot C_{\ell}$. Afterwards only (\ref{eq:cbna_mean}) and (\ref{eq:cbna_variance}) have to be executed, which only induces additional FLOPS in the order of $C_{\ell}$. For the given network architecture, the feature map resolution is up to the order of $H_{\ell}\cdot W_{\ell}\sim 10^6$ (image resolution of $512\times 1024$), while the maximum number of channels is only in the order of $C_{\ell}\sim 10^3$, which is why the main complexity of CBNA is caused by (\ref{eq:cbna_feature_mean}) and (\ref{eq:cbna_feature_variance}). This also explains, why the number of additional FLOPs of CBNA and C-Li/C-Zhang in Table~\ref{tab:runtime_comparison} appears to be equal, as the additional complexity induced by (\ref{eq:cbna_mean}) and (\ref{eq:cbna_variance}) is negligible.

\ifCLASSOPTIONcaptionsoff
  \newpage
\fi

\bibliographystyle{IEEEtran}
\bibliography{IEEEabrv,./bib/ifn_spaml_bibliography}

\vspace{-1cm}
\begin{IEEEbiography}[{\includegraphics[width=1in,height=1.25in,clip,keepaspectratio]{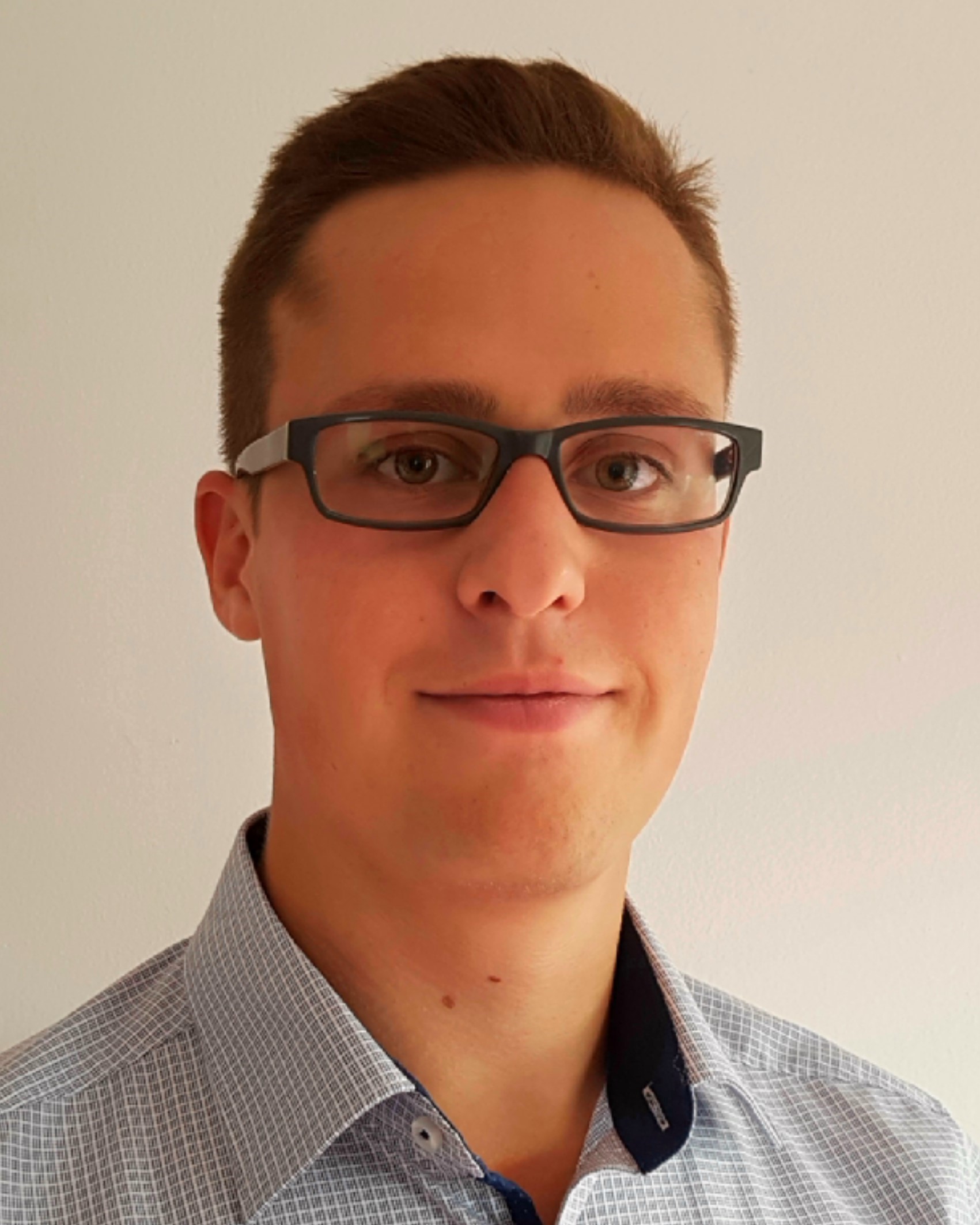}}]{Marvin Klingner}
(m.klingner@tu-bs.de) received a B.Sc.\ degree in 2016 and a M.Sc.\ degree in 2018 in physics from Georg-August-Universität Göttingen, Germany. He is currently a Ph.D.\ student in the Faculty of Electrical Engineering, Information Technology, Physics at Technische Universität Braunschweig, Germany.
His research interests lie in self-supervised 3D geometry perception with neural networks and in multi-task learning and domain adaptation approaches for neural networks with focus on computer vision tasks.
He is the recipient of the Dr. Berliner - Dr. Ungewitter award of the Faculty of Physics at Georg-August-Universität Göttingen, Germany, in 2018, and was given a CVPR Workshop Best Paper Award in 2020 and 2021. 
\end{IEEEbiography}
\vspace{-1cm}

\begin{IEEEbiography}[{\includegraphics[width=1in,height=1.25in,clip,keepaspectratio]{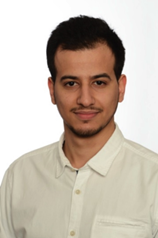}}]{Mouadh Ayache}
(m.ayache@tu-bs.de) recieved a B.Sc.\ degree in 2019 in industrial engineering, specialised in electrical engineering from the Technische Universität Braunschweig, Germany where he currently studies M.Sc. in electrical engineering. Since January 2021, he has been writing his master thesis "Adaptive Online Domain Adaption without Source Data" under the supervision of T. Fingscheidt and M. Klingner with the Signal Processing and Machine Learning Group. He is interested in deep learning applications in autonomous driving, security and in digital hardware design.
\end{IEEEbiography}
\vspace{-1cm}
\begin{IEEEbiography}[{\includegraphics[width=1in,height=1.25in,clip,keepaspectratio]{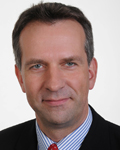}}]{Tim Fingscheidt}
(t.fingscheidt@tu-bs.de) received the Dipl.-Ing.\ degree in electrical engineering in 1993 and the Ph.D.\ degree in 1998 from RWTH Aachen University, Germany.
He joined AT{\&}T Labs, Florham Park, NJ, USA, in 1998, and Siemens AG (Mobile Devices), Munich, Germany, in 1999.
With Siemens Corporate Technology, Munich, Germany, he was leading the speech technology development activities (2005–2006).
Since 2006, he has been a Full Professor with the Institute for Communications Technology, Technische Universität Braunschweig, Germany.
His research interests include speech technology and vision for autonomous driving.
He is the recipient of several awards, including the Vodafone Mobile Communications Foundation prize in 1999 and the 2002 ITG prize of the Association of German Electrical Engineers (VDE ITG).
In 2017 and 2020, he co-authored the ITG award-winning publication, and in 2019, 2020, and 2021 he was given the Best Paper Award of a CVPR Workshop.
He has been the Speaker of the Speech Acoustics Committee ITG AT3 since 2015.
He was an Associate Editor for the IEEE TRANSACTIONS ON AUDIO, SPEECH, AND LANGUAGE PROCESSING (2008–2010), and he has been a member of the IEEE Speech and Language Processing Technical Committee (2011–2018).
\end{IEEEbiography}

\end{document}